\pdfoutput=1
\documentclass[11pt]{article}

\usepackage[final]{acl}

\usepackage{times}
\usepackage{latexsym}

\usepackage[T1]{fontenc}
\usepackage[utf8]{inputenc}

\usepackage{microtype}

\usepackage{inconsolata}

\usepackage{graphicx}

\usepackage{tcolorbox}
\usepackage{xcolor}
\usepackage{listings}
\tcbuselibrary{skins, breakable}

\definecolor{waymogreen}{HTML}{00E89D}
\definecolor{waymolgreen}{HTML}{99F7D7} % alpha=60%
\definecolor{waymollgreen}{HTML}{CCFAEB} % alpha=80%
\definecolor{waymoblue}{HTML}{0077FF}
\definecolor{waymolblue}{HTML}{99B7FF}  % alpha=60%
\definecolor{waymollblue}{HTML}{CCE4FF} % alpha=80%
\definecolor{waymolgray}{HTML}{F0F0F0}  % 浅灰色
\definecolor{mylightgray}{gray}{0.6} % 灰度值

\usepackage{multicol}
\usepackage{multirow}
\usepackage{amsmath}
\usepackage{amssymb}
\usepackage{bbding}
\usepackage{subfigure}
\usepackage{colortbl}
\usepackage{pifont}
\usepackage{bbm}
\usepackage{makecell}
\usepackage{bbding}
\usepackage{mathrsfs}
\usepackage{bm}
\usepackage{CJKutf8}
\usepackage{booktabs}
\usepackage{tabu}
\usepackage{fontawesome}

\usepackage{booktabs}   % 表格线
\usepackage{colortbl}   % 表格颜色
\usepackage{xcolor}     % 颜色定义
\usepackage{arydshln}   % 虚线支持
\usepackage{amsmath} % 确保导包
\definecolor{aliceblue}{rgb}{0.94, 0.97, 1.0}
\definecolor{ClosedColor}{HTML}{D8EEF2}  % 你也可以换成 cyan!15 或其他
\definecolor{OpenColor}{HTML}{FDEBDD}    % 你也可以换成 red!15 或其他
\definecolor{HeaderColor}{gray}{.85}
\usepackage{colortbl}
\arrayrulecolor{black}    % 恢复默认的表格线颜色

\usepackage{algorithm}
\usepackage{algpseudocode}
\title{From Imitation to Discrimination: Progressive Curriculum Learning for Robust Web Navigation}

\author{
  {\bf Chuang Peng}\textsuperscript{\rm 1*},
  {\bf Wei Zhang}\textsuperscript{\rm 2}\thanks{\ Equal contribution.},
  {\bf Renshuai Tao}\textsuperscript{\rm 1}\thanks{\ Corresponding author.},
  {\bf Xinhao Zhang}\textsuperscript{\rm 1},
  {\bf Jian Yang}\textsuperscript{\rm 2}\\
   \textsuperscript{\rm 1}Beijing Jiaotong University;
   \textsuperscript{\rm 2}Beihang University;\\
   \texttt{pchuang@bjtu.edu.cn, zwpride@buaa.edu.cn, rstao@bjtu.edu.cn}
}

\begin{document}
\maketitle
\begin{abstract}
Text-based web agents offer computational efficiency for autonomous web navigation, yet developing robust agents remains challenging due to the noisy and heterogeneous nature of real-world HTML. Standard Supervised Fine-Tuning (SFT) approaches fail in two critical dimensions: they lack discrimination capabilities to reject plausible but incorrect elements in densely populated pages, and exhibit limited generalization to unseen website layouts. To address these challenges, we introduce the \textbf{Triton} dataset (590k instances) and a progressive training curriculum. Triton is constructed via Structural-Semantic Hard Negative Mining, which explicitly mines topologically similar distractors, and a Dual-Agent Consensus pipeline that synthesizes diverse cross-domain tasks with strict verification. Building upon this foundation, our progressive curriculum produces three models: \textbf{Triton-SFT-32B} for basic imitation, \textbf{Triton-ORPO-32B} for robust discrimination via Odds Ratio Preference Optimization, and \textbf{Triton-GRPO-32B} for long-horizon consistency through Group Relative Policy Optimization. Empirical evaluation on Mind2Web demonstrates that \textbf{Triton-GRPO-32B} achieves state-of-the-art performance among open-source models with 58.7\% Step Success Rate, surpassing GPT-4.5 (42.4\%) and Claude-4.5 (41.4\%) by over 16\%, validating that specialized data curriculum outweighs raw parameter scale for web navigation.
\end{abstract}

\section{Introduction}

Enabling autonomous agents to navigate real-world websites through natural language instructions is a cornerstone of general-purpose artificial intelligence~\cite{deng2023mind2web,zhou2023webarena,yang2025code}. While recent text-based web agents offer computational efficiency over multimodal counterparts~\cite{zheng2024seeact,hong2024cogagent}, developing robust agents remains formidable due to two critical challenges: \textbf{(1) Lack of Discrimination}: Standard Supervised Fine-Tuning (SFT) teaches agents ``what to do'' but fails to train ``what \textit{not} to do,'' leading to hallucinations where models select plausible but incorrect elements in densely populated pages. \textbf{(2) Limited Generalization}: Relying solely on domain-specific training data constrains the agent's ability to ground instructions in unseen website layouts and diverse DOM structures.

%%%%%%%%%
\begin{figure}[!t]
  \centering
\includegraphics[width=1\linewidth]{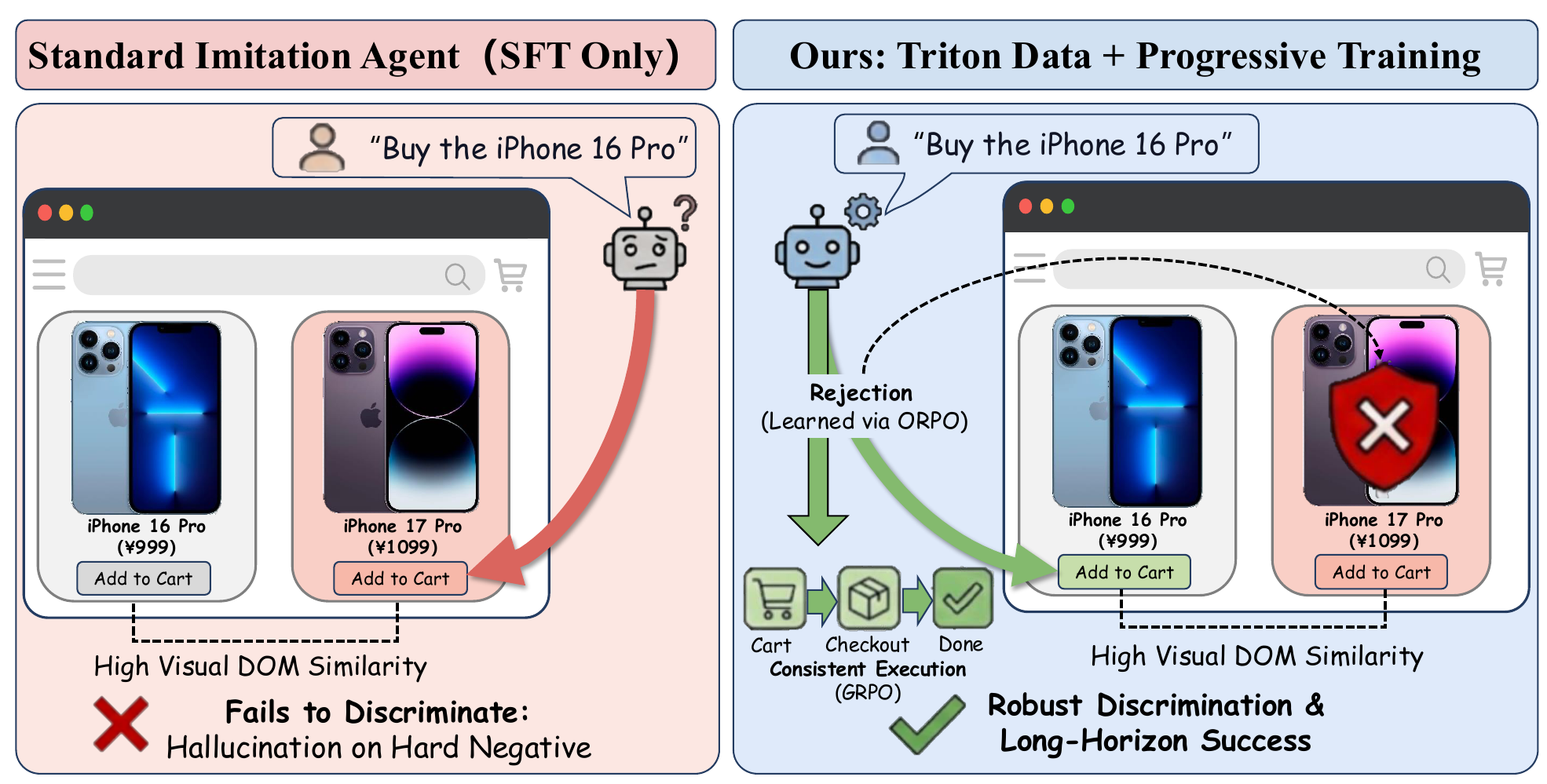}
    \vspace{-20pt}
    \caption{Standard imitation learning hallucinates on structurally similar elements (incorrectly selecting iPhone 17 Pro when instructed to buy iPhone 16 Pro), while Triton trained with progressive curriculum (ORPO for discrimination + GRPO for consistency) achieves target identification and multi-step task completion.}
    \label{fig:1}
   \vspace{-22pt}
\end{figure}

To address these limitations, we introduce the \textbf{Triton dataset} and a \textbf{progressive training curriculum}. For \textbf{discrimination}, Triton is constructed via \textit{Structural-Semantic Hard Negative Mining}, which calculates DOM-tree edit distances to mine structurally identical distractors that force fine-grained attribute reasoning. We further introduce counterfactual ``rejection'' samples to curb hallucinations. For \textbf{generalization}, we employ a \textit{Dual-Agent Consensus} pipeline that synthesizes 290k diverse cross-domain navigation tasks from WebSight~\cite{laurent2024websight}, where a Generator creates instructions across three cognitive levels and a Verifier ensures strict grounding alignment. Building upon this 590k instance dataset, our progressive curriculum evolves models through three stages: \textbf{Triton-SFT-32B} establishes foundational instruction-following via supervised fine-tuning; \textbf{Triton-ORPO-32B} sharpens decision boundaries through Odds Ratio Preference Optimization~\cite{hong2024orpo}, penalizing high-probability errors without requiring a separate reward model; \textbf{Triton-GRPO-32B} enhances long-horizon consistency via Group Relative Policy Optimization~\cite{shao2024deepseekmath}, sampling groups of outputs and rewarding outcome consistency to stabilize multi-step execution.

% Building upon this rigorous data foundation, we introduce a \textbf{Progressive Training Curriculum} that transcends standard SFT. Our pipeline consists of three stages: (1) \textbf{Foundation via SFT}: We first establish the agent's instruction-following capability using our full Triton dataset. (2) \textbf{Discrimination via ORPO}: To sharpen decision boundaries, we employ Odds Ratio Preference Optimization (ORPO)~\cite{hong2024orpo}. By constructing "Winner-Loser" pairs from the model's own high-probability errors, we explicitly penalize hallucinations without the need for a separate reward model. (3) \textbf{Consistency via GRPO}: Finally, we apply Group Relative Policy Optimization (GRPO)~\cite{shao2024deepseekmath} on a subset of complex tasks. By sampling groups of outputs and rewarding outcome consistency, we further stabilize the agent's performance in long-horizon scenarios.
% Empirical results on the Mind2Web benchmark demonstrate that our framework achieves state-of-the-art performance among open-source models, significantly outperforming baselines in both in-domain and cross-domain settings.
% Our work highlights that a principled data curriculum—moving from discrimination to consistency—is key to unlocking the full potential of web agents.

Our \textbf{key contributions} are threefold:

\begin{itemize}
    \item \textbf{Data Innovation}: We introduce \textbf{Triton}, a 590k-instance dataset featuring: (1) topological hard negatives mined via DOM-tree edit distance to enforce fine-grained structural reasoning; (2) counterfactual rejection samples to train explicit decision boundaries; and (3) 290k synthetic cross-domain tasks constructed via a Dual-Agent Consensus pipeline, achieving 96\% human-verified quality. This addresses both discrimination and generalization challenges.
    
    \item \textbf{Methodological Contribution}: We propose a \textbf{progressive training curriculum} that systematically evolves models through three stages: \textbf{Triton-SFT-32B} for foundational imitation, \textbf{Triton-ORPO-32B} for discriminative optimization by penalizing high-probability errors, and \textbf{Triton-GRPO-32B} for long-horizon consistency via group-based policy optimization. This demonstrates that discrimination and consistency require specialized alignment beyond standard behavioral cloning.
    
    \item \textbf{Empirical Validation}: \textbf{Triton-GRPO-32B} achieves state-of-the-art performance among open-source models (58.7\% Step SR on Mind2Web), surpassing GPT-4.5 (42.4\%) and Claude-4.5 (41.4\%) by over 16 points. Despite having only 32B parameters, it more than doubles the performance of 671B DeepSeek-V3 (25.2\%), validating that specialized data curriculum outweighs raw parameter scale. Ablation studies confirm additive gains: SFT (47.6\%) + ORPO (+5.6\%) + GRPO (+5.5\%).
\end{itemize}

\begin{figure*}[t!]
\begin{center}
    \includegraphics[width=1.0\textwidth]{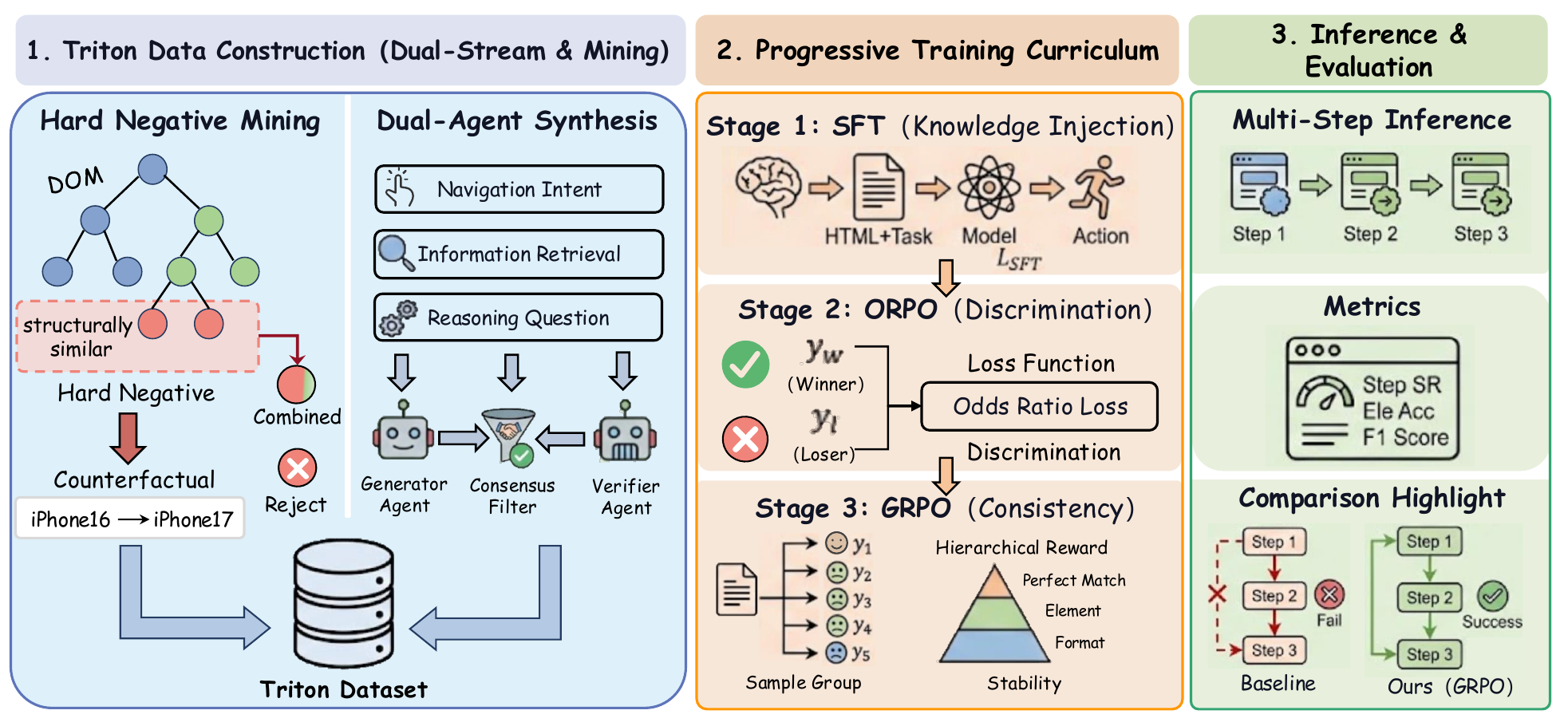}
    \vspace{-22pt}
    \caption{Overview of our proposed framework. (Left) We construct the Triton dataset by mining topological hard negatives and synthesizing cross-domain tasks via Dual-Agent Consensus. (Middle) This data fuels a progressive curriculum that evolves the model from imitation (SFT) to discrimination (ORPO) and consistency (GRPO), (Right) enabling robust execution in multi-step environments.}
    \label{fig:framework}
    \vspace{-22pt}
\end{center}
\end{figure*}

\section{The Triton Dataset}
\label{sec:dataset}

To overcome the lack of discriminative signals and limited layout diversity in existing datasets, we introduce Triton, a dataset of 590k instances. It is constructed via a two-stage pipeline: \textit{Discriminative Trajectory Mining} on in-domain data and \textit{Synthetic Visual-Structural Grounding} on cross-domain layouts.

\subsection{Discriminative Trajectory Enhancement}
\label{sec:data_mining}
Standard instruction tuning often treats web navigation as a simple sequence generation task, ignoring the complex decision boundaries inherent in noisy DOM environments. To equip the agent with the ability to distinguish correct elements from plausible distractors, we refine the Mind2Web training set through two strategies. We expand the Mind2Web training set from 7k trajectories to 300k training instances, where each trajectory is augmented with 20 hard negatives and counterfactual samples, yielding approximately a 40× expansion.

\paragraph{Topological Hard Negative Mining.}
Randomly sampling negative elements from the DOM tree is inefficient, as most elements are easily distinguishable. To mine \textit{hard negatives} elements, that are structurally or semantically similar to the target, we employ a hybrid similarity metric. 
Given a ground-truth element $e^+$, we calculate its similarity with candidate $e^-$ based on two factors: (1) \textbf{DOM-Tree Edit Distance (TED)}, which measures the topological proximity (e.g., adjacent items in a list), and (2) \textbf{Attribute Jaccard Similarity}, which measures the overlap of textual attributes (e.g., class names, aria-labels). 
We select the top-$K$ elements (where $K=20$) with the highest combined similarity scores as hard negatives. This forces the model to attend to fine-grained structural and attribute differences rather than relying on superficial heuristics.

\paragraph{Counterfactual Trajectory Generation.}
To address the hallucination problem where agents execute actions even when the target is absent, we generate counterfactual "Rejection" samples via \textit{Instruction Perturbation}. We apply two specific perturbation strategies:
(1) \textbf{Entity Swap}: Replacing the target entity in the instruction with a similar but absent entity (e.g., changing "Buy iPhone 16" to "Buy iPhone 17" when only the former exists).
(2) \textbf{Action Mismatch}: Modifying the requested action type to be incompatible with the available elements (e.g., "Click Register" on a page with only a "Login" button).
For these perturbed instructions, the model is trained to output a special \texttt{None} token, explicitly learning decision boundaries for rejection.

\subsection{Synthetic Visual-Structural Grounding}
\label{sec:data_synthesis}
Relying solely on Mind2Web limits the agent's generalization to unseen website layouts. To bridge this gap, we leverage the WebSight dataset~\cite{laurent2024websight}, which contains diverse HTML codes and screenshots, to synthesize a large-scale visual-grounding corpus.

\paragraph{Diversity-Driven Instruction Synthesis.}
We employ \textbf{Qwen3-Coder-480B-A35B-Instruct}~\cite{qwen3coder} as the generator to synthesize natural language instructions based on the HTML snippets. To ensure task diversity, we design three distinct prompt templates covering different cognitive levels: \textbf{1) Navigation Intent}: Direct operational commands (e.g., \textit{"Click the login button at the top right."}). \textbf{2) Information Retrieval}: Queries requiring content identification (e.g., \textit{"Find the price of the item listed."}). \textbf{3) Reasoning Question}: Tasks requiring logic deduction (e.g., \textit{"Which button should I click to proceed to checkout?"}).

\paragraph{Dual-Agent Consensus Filtering.}
Automated synthesis often suffers from hallucinations or incorrect grounding. To ensure data quality, we introduce a \textbf{Dual-Agent Consensus} pipeline. 
While Qwen3-Coder-480B-A35B-Instruct generates the (Instruction, Element) pairs, we utilize the more powerful \textbf{Qwen3-32B-Instruct}~\cite{qwen3} as a \textit{Verifier}. The Verifier receives the synthesized instruction and the raw HTML, and independently predicts the target element. A data sample is retained only if the Verifier's prediction strictly matches the Generator's label (Exact Match) or shares a highly overlapping DOM path. This strict filtering process yields 290k high-fidelity samples from an initial pool of over 500k.

\paragraph{Pilot Human Verification.}
To validate the reliability of our automated pipeline, we conducted a pilot study on a randomly sampled subset of 200 synthesized instances. Human annotators verified the correctness of the instruction-element pairs. The study revealed a \textbf{96\% pass rate}, confirming that our Dual-Agent Consensus mechanism effectively filters out noise and produces training data comparable to human-annotated quality.

\subsection{Data Statistics and Analysis}
\label{sec:data_stats}
\begin{figure}[t]
    \centering
    \includegraphics[width=\linewidth]{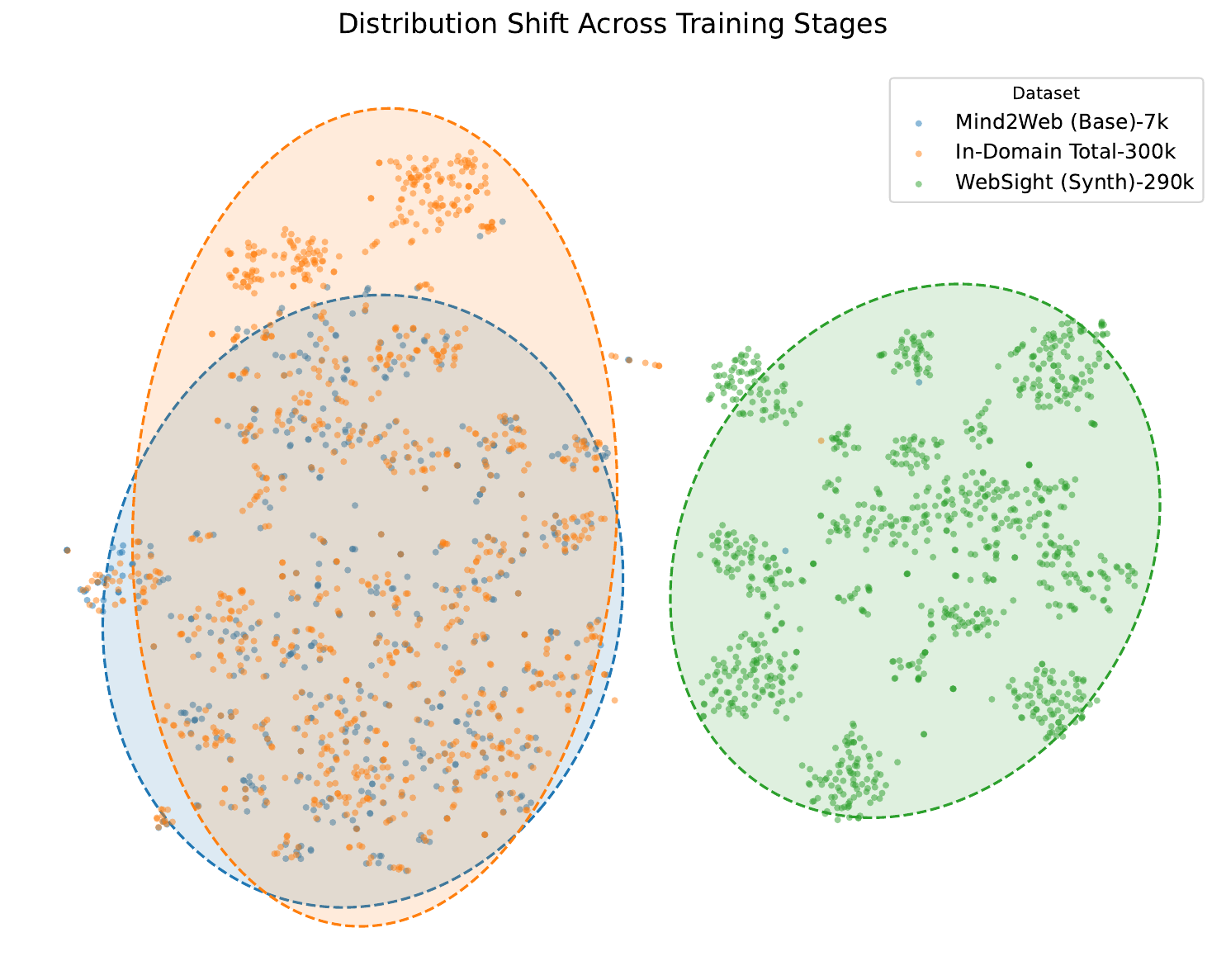}
    \vspace{-22pt}
    \caption{Visualization of semantic diversity via t-SNE. While the original Mind2Web data (Blue) forms dense, domain-specific clusters, our synthetic WebSight data (Green) spans a broader semantic space. This complementary distribution demonstrates that Triton effectively fills the "generalization gaps" left by standard behavioral cloning datasets.}
    \label{fig:data_diversity}
    \vspace{-22pt}
\end{figure}

As detailed in Table ~\ref{tab:triton_composition}, Triton scales to a total of 590k instances through a hybrid construction strategy that balances discriminative depth with semantic breadth. The in-domain component originates from Mind2Web’s limited 7k trajectories but is substantially augmented into 300k training samples. This expansion is driven by our DOM-Tree Mining (140k) and Counterfactual Perturbation (153k) techniques, which transform successful trajectories into hard-negative discrimination tasks.
To assess the semantic coverage, we visualize the instruction embeddings via t-SNE in Figure ~\ref{fig:data_diversity}. The distribution reveals a complementary relationship: while Mind2Web (Blue) forms dense clusters representing deep, task-specific logic within limited domains, our synthetic WebSight data Green) occupies a significantly broader region of the semantic space. This wide coverage serves as a structural regularizer, bridging the gaps between specific website layouts and equipping the agent with general-purpose navigation capabilities required for unseen domains.

\begin{table}[t]
    \centering
    \small
    \setlength{\tabcolsep}{5pt}
    \begin{tabular}{lcc}
        \toprule
        \textbf{Component} & \textbf{Method} & \textbf{\# Samples} \\
        \midrule
        Mind2Web (Base) & - & 7k \\
        \quad + Hard Neg. & DOM-Tree Mining & 140k \\
        \quad + Rejection & Perturbation & 153k \\
        \cmidrule(lr){1-3}
        \textit{In-Domain Total} & & \textit{300k} \\
        \cmidrule(lr){1-3}
        WebSight (Synth.) & Dual-Agent & 290k \\
        \midrule
        \textbf{Triton Total} & & \textbf{590k} \\
        \bottomrule
    \end{tabular}
    \caption{Composition of the Triton Dataset}
    \label{tab:triton_composition}
    \vspace{-15pt}
\end{table}

\section{Progressive Training Curriculum}
\label{sec:training}

While high-quality data is provided, effectively internalizing this knowledge requires a strategic training curriculum. Standard behavioral cloning alone is insufficient for web agents to master discrimination and precision. Therefore, we propose a \textbf{Progressive Alignment Curriculum} consisting of three stages: (1) \textit{Foundation via SFT} to establish basic instruction following; (2) \textit{Discrimination via ORPO} to sharpen decision boundaries against hard negatives; (3) \textit{Consistency via GRPO} to refine action precision in complex scenarios.

\subsection{Stage I: Foundation via SFT}
In the first stage, we treat web navigation as a sequence modeling task. We utilize the full Triton dataset ($\mathcal{D}_{\text{all}} \approx 590k$) to perform Supervised Fine-Tuning (SFT). The goal is to maximize the likelihood of the ground-truth action $y$ given the HTML observation $x$.
\begin{equation}
    \mathcal{L}_{\text{SFT}} = - \mathbb{E}_{(x, y) \sim \mathcal{D}_{\text{all}}} \left[ \log P_\theta(y | x) \right]
\end{equation}
This stage functions as a "Knowledge Injection" phase, ensuring the agent learns the fundamental syntax of HTML and the diverse intent patterns introduced by our Multi-Task formatting.

\subsection{Stage II: Discrimination via ORPO}
SFT models often suffer from the "confusion" problem—assigning high probabilities to incorrect elements that are semantically similar to the target (i.e., Hard Negatives). To address this, we employ Odds Ratio Preference Optimization (ORPO)~\cite{hong2024orpo}, which penalizes the generation of incorrect answers without requiring a separate reward model.

\paragraph{Self-Synthesized Preference Pairs.}
Instead of reusing static negative samples, we construct dynamic preference pairs based on the SFT model's own errors. We perform on-policy sampling on the training set, generating $N=5$ outputs for each instruction. We construct pairs $(y_w, y_l)$ where: \textbf{Winner ($y_w$)}: The ground-truth trajectory. \textbf{Loser ($y_l$)}: An incorrect trajectory generated by the model. This represents the model's current ``blind spots.''

This results in a discriminative subset $\mathcal{D}$. We define the odds of generating a sequence $y$ as $g(y|x) = \frac{P(y|x)}{1-P(y|x)}$. The ORPO objective optimizes the log odds ratio of the winner over the loser:
\begin{equation}
    \small
    \mathcal{L}_{\text{ORPO}} = \mathcal{L}_{\text{SFT}} - \lambda \mathbb{E}_{\mathcal{D}} \left[ \log \sigma \left( \log \frac{g(y_w|x)}{g(y_l|x)} \right) \right]
\end{equation}
Crucially, although the underlying data source overlaps with SFT, the \textit{learning objective} shifts from pure imitation to discrimination, allowing the model to learn from its own mistakes.

\subsection{Stage III: Consistency via GRPO}
In the final stage, we aim to enhance the agent's stability and precision, particularly for complex, long-horizon tasks. We employ Group Relative Policy Optimization (GRPO)~\cite{shao2024deepseekmath} on a curated subset of 80k complex instances. Unlike PPO, GRPO eliminates the need for a critic model by normalizing advantages within a group of sampled outputs.

\paragraph{Hierarchical Reward Shaping.}
Since our task involves strict format constraints and requires high precision in text generation (e.g., typing exact strings), a sparse binary reward (0/1) provides insufficient supervision. We design a \textit{Hierarchical Reward Function} that guides the agent step-by-step:
\begin{equation}
    R(y) = R_{\text{fmt}} + \mathbb{I}_{\text{opt}} \cdot (R_{\text{opt}} + R_{\text{F1}} + R_{\text{perf}})
\end{equation}
where: $R_{\text{fmt}}$ (0.1): A small reward for adhering to the valid output format (valid Element ID and Operation fields). $\mathbb{I}_{\text{opt}}$: An indicator function that is 1 if the selected option (e.g., ``Option C'') matches the ground truth, and 0 otherwise. This acts as a gatekeeper; if the option is wrong, subsequent rewards are zeroed out to prevent reward hacking. $R_{\text{opt}}$ (1.0): A major reward for selecting the correct element. $R_{\text{F1}}$ (0.0-1.0): The token-level F1 score between the predicted action string and the ground truth. This provides dense feedback for "almost correct" actions. $R_{\text{perf}}$ (1.0): A bonus reward awarded only for a perfect match (F1=1.0).

During training, for each input $x$, we sample a group of $G=5$ outputs $\{y_1, ..., y_G\}$ and optimize the policy to maximize the expected reward relative to the group baseline. This mechanism encourages the model to converge towards the most precise and consistent action execution path.

% Table Definition
\begin{table}[t]
    \centering
    \resizebox{\columnwidth}{!}{%
        \setlength{\tabcolsep}{4.5pt} % 保持你原有的列间距设置
        \begin{tabular}{lcccc}
        \toprule
        \textbf{Split} & \textbf{\# Tasks} & \textbf{\# Domains} & \textbf{\# Websites} & \textbf{Avg \# Actions} \\
        \midrule
        Train & 1,009 & 17 & 73 & 7.7 \\
        Cross-Task & 177 & 17 & 64 & 7.6 \\
        Cross-Website & 142 & 9 & 10 & 7.2 \\
        Cross-Domain & 694 & 13 & 53 & 5.9 \\
        \bottomrule
        \end{tabular}%
    }
    \caption{Statistics of the \textsc{Mind2Web} dataset. The benchmark is divided into three test splits to evaluate generalization difficulty. Note that the Cross-Domain split contains the largest number of tasks but the shortest average trajectory length.}
    \label{tab:mind2web_stats}
    \vspace{-15pt}
\end{table}

%%%%%%%%%%%%%%图表
\begin{table*}[t!]
  \centering
  \small
  \renewcommand{\arraystretch}{1.1}
  \resizebox{\textwidth}{!}{%
    \begin{tabular}{ll|cc|cccccc|cccccc|cccccc|}
      \toprule
       \multirow{2}{*}{Model} & \multirow{2}{*}{Size} & \multicolumn{2}{c}{All} & \multicolumn{6}{c}{Cross-Domain} & \multicolumn{6}{c}{Cross-Task} & \multicolumn{6}{c}{Cross-Website} \\
        & &Score& SR & Score &SR & Ele & Op & M.Ele & M.Op 
       & Score &SR & Ele & Op & M.Ele & M.Op 
        & Score &SR & Ele & Op & M.Ele & M.Op \\
      \midrule
      \rowcolor{HeaderColor}\multicolumn{22}{c}{\textit{Closed-Source LLMs}} \\
      \midrule
      \rowcolor{ClosedColor} \small Claude-4-5-noThink & \faLock{} & 58.8 &41.4  &58.1 &41.8 &48.3 &67.0 &49.9 &67.2 &61.5 &43.7 &52.0 &68.3 &55.2 &70.4 & 56.9& 38.7 & 46.8&65.1 &48.7 &67.0  \\
      \rowcolor{ClosedColor} \small Gemini-2.0-Flash & \faLock{} & 53.1 & 35.0 & 52.6 & 35.2 & 43.1 & 61.2 & 44.5 & 61.5 & 55.7 & 37.1 & 46.5 & 63.4 & 48.8 & 64.2 & 51.0 & 32.8 & 41.2 & 59.5 & 43.1 & 60.2 \\
      \rowcolor{ClosedColor} \small Gemini-2.5-Flash-preview-04-17 & \faLock{} & 55.3 & 38.0 & 54.9 & 38.4 & 45.4 & 63.5 & 46.8 & 63.9 & 57.7 & 40.1 & 48.2 & 65.1 & 50.5 & 66.8 & 53.3 & 35.6 & 43.5 & 61.8 & 45.2 & 62.5 \\
      \rowcolor{ClosedColor} \small Gemini-2.5-pro-preview-05-06 & \faLock{} & 59.1 & 41.5 & 58.3 & 41.6 & 48.5 & 67.2 & 50.1 & 67.5 & 61.7 & 43.9 & 52.2 & 68.5 & 55.4 & 70.6 & 57.2 & 39.1 & 47.1 & 65.4 & 49.0 & 67.2 \\
      \rowcolor{ClosedColor} \small GPT-4.1-mini-2025-04-14 & \faLock{} & 52.0 & 33.2 & 51.6 & 33.5 & 42.5 & 59.8 & 43.8 & 60.2 & 54.3 & 35.2 & 45.2 & 61.5 & 47.5 & 62.8 & 50.1 & 31.0 & 40.5 & 58.2 & 42.1 & 59.5 \\
      \rowcolor{ClosedColor} \small GPT-4.1-nano-2025-04-14 & \faLock{} & 48.4 & 28.4 & 47.8 & 28.5 & 39.5 & 55.2 & 40.8 & 55.8 & 50.7 & 30.2 & 42.1 & 57.5 & 44.2 & 58.9 & 46.7 & 26.5 & 37.8 & 54.1 & 39.2 & 55.5 \\
      \rowcolor{ClosedColor} \small GPT-4.5-preview-2025-02-27 & \faLock{} & 59.8 & 42.4 & 59.0 & 42.5 & 49.2 & 67.8 & 50.8 & 68.1 & 62.4 & 44.6 & 52.8 & 69.2 & 56.1 & 71.3 & 58.0 & 40.2 & 47.9 & 66.0 & 49.8 & 68.1 \\
      \rowcolor{ClosedColor} \small Grok-3 & \faLock{} & 58.5 & 41.1 & 57.8 & 41.2 & 47.9 & 66.8 & 49.5 & 67.0 & 61.1 & 43.5 & 51.5 & 68.0 & 54.8 & 70.1 & 56.7 & 38.5 & 46.5 & 64.9 & 48.4 & 66.8 \\
      \rowcolor{ClosedColor} \small Grok-3-fast & \faLock{} & 54.2 & 36.6 & 53.7 & 36.8 & 44.2 & 62.1 & 45.8 & 62.5 & 56.7 & 38.9 & 47.5 & 64.2 & 49.6 & 65.5 & 52.2 & 34.2 & 42.8 & 60.5 & 44.1 & 61.2 \\
      \rowcolor{ClosedColor} \small o4-mini-2025-04-16 & \faLock{} & 56.8 & 39.2 & 56.1 & 39.2 & 46.8 & 64.5 & 48.2 & 64.9 & 59.4 & 41.5 & 50.1 & 66.8 & 52.5 & 68.2 & 54.9 & 36.8 & 45.2 & 63.1 & 46.8 & 64.5 \\
      \midrule
      \rowcolor{HeaderColor}\multicolumn{22}{c}{\textit{Open-Source LLMs}} \\
      \midrule
      \rowcolor{OpenColor} \small Qwen3-0.6B-Instruct Think & 0.6B & 3.5 & 0.0 & 4.8 & 0.0 & 6.0 & 0.0 & 7.3 & 0.0 & 3.0 & 0.0 & 5.0 & 0.0 & 7.0 & 0.0 & 2.6 & 0.0 & 4.4 & 0.0 & 5.8 & 0.0 \\
      \rowcolor{OpenColor} \small Qwen3-0.6B-Instruct Chat & 0.6B & 4.2 & 0.3 & 4.3 & 0.3 & 0.4 & 7.9 & 0.5 & 8.3 & 4.5 & 0.3 & 0.4 & 8.8 & 0.4 & 8.3 & 3.8 & 0.4 & 0.7 & 6.5 & 1.0 & 6.8 \\
      \rowcolor{OpenColor} \small Qwen3-1.7B-Instruct Chat & 1.7B & 23.9 & 6.4 & 19.0 & 7.3 & 10.0 & 43.7 & 12.3 & 44.1 & 28.9 & 6.5 & 9.3 & 46.8 & 12.8 & 46.6 & 23.9 & 5.4 & 8.7 & 37.8 & 10.9 & 38.1 \\
      \rowcolor{OpenColor} \small Qwen3-1.7B-Instruct Think & 1.7B & 4.8 & 0.1 & 5.3 & 0.2 & 6.0 & 2.7 & 6.6 & 2.7 & 5.3 & 0.1 & 2.3 & 4.1 & 6.1 & 2.6 & 7.2 & 0.1 & 2.5 & 0.1 & 7.7 & 2.3 \\
      \rowcolor{OpenColor} \small Qwen3-4B-Instruct Chat & 4B & 29.7 & 11.2 & 23.6 & 12.5 & 14.4 & 49.0 & 16.7 & 50.4 & 33.2 & 11.3 & 13.1 & 50.9 & 16.5 & 52.4 & 32.3 & 10.1 & 14.4 & 48.3 & 17.6 & 49.0 \\
      \rowcolor{OpenColor} \small Qwen3-4B-Instruct Think & 4B & 2.2 & 0.6 & 1.8 & 0.6 & 1.1 & 3.7 & 1.3 & 3.5 & 2.4 & 0.8 & 1.4 & 3.6 & 1.2 & 3.5 & 2.4 & 0.5 & 1.2 & 3.5 & 1.2 & 3.8 \\
      \rowcolor{OpenColor} \small Qwen3-8B-Instruct Think & 8B & 9.6 & 0.9 & 8.0 & 1.0 & 5.9 & 14.1 & 6.1 & 14.3 & 10.6 & 0.8 & 6.2 & 14.2 & 7.5 & 14.6 & 10.2 & 1.0 & 6.0 & 13.5 & 7.1 & 14.0 \\
      \rowcolor{OpenColor} \small Qwen3-8B-Instruct Chat & 8B & 18.5 & 5.9 & 17.5 & 5.9 & 16.2 & 19.0 & 18.7 & 18.2 & 19.8 & 6.0 & 15.3 & 22.5 & 19.4 & 21.8 & 18.1 & 5.8 & 14.2 & 21.1 & 17.3 & 19.8 \\
      \rowcolor{OpenColor} \small Qwen3-14B-Instruct Chat & 14B & 16.2 & 1.6 & 14.9 & 1.9 & 12.5 & 20.1 & 14.3 & 21.2 & 17.0 & 1.3 & 11.7 & 20.5 & 14.1 & 21.5 & 16.9 & 1.7 & 11.2 & 20.5 & 13.6 & 22.3 \\
      \rowcolor{OpenColor} \small Qwen3-14B-Instruct Think & 14B & 0.3 & 0.1 & 0.4 & 0.1 & 0.3 & 0.8 & 0.3 & 0.7 & 0.3 & 0.0 & 0.1 & 0.5 & 0.1 & 0.4 & 0.3 & 0.0 & 0.1 & 0.5 & 0.1 & 0.5 \\
      \rowcolor{OpenColor} \small Qwen3-32B-Instruct Think & 32B & 1.8 & 0.2 & 1.6 & 0.2 & 1.1 & 2.7 & 1.4 & 3.0 & 2.4 & 0.1 & 1.4 & 2.8 & 1.8 & 3.4 & 1.6 & 0.2 & 0.6 & 2.3 & 0.7 & 2.7 \\
      \rowcolor{OpenColor} \small Qwen3-32B-Instruct Chat & 32B & 36.9 & 17.7 & 33.0 & 17.9 & 26.5 & 49.2 & 29.9 & 48.1 & 40.3 & 17.9 & 25.6 & 53.5 & 30.3 & 51.8 & 37.2 & 17.3 & 24.1 & 48.7 & 26.6 & 49.5 \\
      \rowcolor{OpenColor} \small Qwen2.5-Coder-14B-Instruct & 14B & 33.7 & 16.2 & 32.0 & 16.1 & 29.0 & 37.4 & 32.5 & 37.3 & 36.3 & 17.2 & 28.6 & 42.4 & 32.1 & 42.1 & 32.9 & 15.2 & 26.4 & 37.7 & 28.9 & 38.6 \\
      \rowcolor{OpenColor} \small Qwen2.5-Coder-32B-Instruct & 32B & 28.7 & 14.6 & 30.1 & 15.5 & 23.7 & 34.6 & 26.0 & 36.3 & 29.8 & 16.1 & 24.0 & 33.4 & 26.7 & 35.5 & 26.2 & 27.9 & 23.1 & 13.1 & 27.9 & 13.1 \\
      \rowcolor{OpenColor} \small Qwen3-Coder-30B-A3B-Instruct & 30B & 32.7 & 7.5 & 36.0 & 7.5 & 41.0 & 19.7 & 42.2 & 20.2 & 30.8 & 6.5 & 42.7 & 15.9 & 45.7 & 18.9 & 31.4 & 8.4 & 40.4 & 19.6 & 44.3 & 21.1 \\
      \rowcolor{OpenColor} \small Qwen3-Next-80B-A3B-Instruct & 80B & 27.9 & 15.2 & 25.9 & 15.6 & 22.6 & 34.3 & 24.3 & 35.0 & 29.9 & 15.7 & 22.5 & 35.5 & 25.0 & 36.6 & 27.8 & 14.3 & 21.0 & 33.2 & 23.0 & 34.1 \\
      \rowcolor{OpenColor} \small Qwen3-Next-80B-A3B-Think & 80B & 17.8 & 5.2 & 16.8 & 5.9 & 13.2 & 26.2 & 14.7 & 28.2 & 19.4 & 5.6 & 12.0 & 25.7 & 13.3 & 26.5 & 17.2 & 4.1 & 10.3 & 22.2 & 12.1 & 24.1 \\
      \rowcolor{OpenColor} \small Qwen3-Coder-480B-A35B-Instruct & 480B & 38.2 & 21.8 & 34.8 & 22.9 & 29.9 & 47.1 & 32.2 & 48.4 & 40.4 & 21.9 & 29.0 & 48.6 & 32.2 & 51.6 & 39.4 & 20.7 & 27.7 & 48.3 & 30.8 & 50.7 \\
      \rowcolor{OpenColor} \small Qwen3-235B-A22B-Think-2507 & 235B & 27.1 & 3.9 & 26.9 & 4.3 & 25.1 & 29.2 & 28.1 & 30.8 & 27.8 & 3.5 & 24.2 & 28.3 & 28.1 & 30.4 & 26.7 & 4.0 & 22.0 & 27.9 & 25.8 & 31.1 \\
      \rowcolor{OpenColor} \small GPT-Oss-20b & 20B & 13.0 & 0.8 & 16.3 & 0.9 & 19.6 & 3.2 & 22.7 & 3.1 & 11.6 & 1.1 & 17.0 & 4.8 & 20.7 & 4.0 & 11.0 & 0.6 & 16.7 & 3.6 & 20.2 & 3.5 \\
      \rowcolor{OpenColor} \small GPT-Oss-120b & 120B & 9.6 & 1.1 & 9.5 & 1.2 & 8.7 & 11.2 & 9.3 & 11.1 & 9.9 & 0.7 & 8.3 & 11.2 & 8.9 & 11.1 & 9.6 & 1.3 & 7.2 & 11.5 & 8.1 & 11.4 \\
      \rowcolor{OpenColor} \small OpenCoder-8B-Instruct & 8B & 13.4 & 4.9 & 12.9 & 4.8 & 19.0 & 7.7 & 17.1 & 7.8 & 15.9 & 5.9 & 21.9 & 9.7 & 21.9 & 10.2 & 11.5 & 4.0 & 14.3 & 8.8 & 13.4 & 9.4 \\
      \rowcolor{OpenColor} \small Llama3.1-8B-Instruct & 8B & 23.4 & 8.3 & 25.6 & 9.2 & 28.1 & 22.2 & 28.3 & 23.7 & 26.1 & 8.5 & 31.4 & 21.0 & 29.2 & 22.8 & 18.6 & 7.1 & 21.6 & 15.0 & 22.0 & 15.9 \\
      \rowcolor{OpenColor} \small DeepSeek-V3 & 37/671B & 43.4 & 25.2 & 38.5 & 26.3 & 30.7 & 59.2 & 33.4 & 60.9 & 47.2 & 25.5 & 30.3 & 60.8 & 33.9 & 63.8 & 44.5 & 23.9 & 29.9 & 57.6 & 32.4 & 58.2 \\
      \rowcolor{OpenColor} \small DeepSeek-R1 & 37/671B & 15.7 & 4.2 & 18.4 & 4.2 & 20.1 & 10.0 & 23.2 & 11.6 & 15.4 & 4.8 & 18.2 & 10.4 & 20.8 & 12.0 & 13.3 & 3.4 & 16.8 & 8.5 & 18.7 & 9.3 \\
      \rowcolor{OpenColor} \small DeepSeek-V3.1-Think &37/671B  & 43.5 & 26.3 & 39.2 & 27.5 & 31.8 & 58.3 & 34.9 & 60.5 & 47.0 & 26.7 & 31.2 & 59.7 & 34.2 & 62.7 & 44.5 & 24.7 & 29.9 & 57.0 & 31.9 & 59.1 \\
      \rowcolor{OpenColor} \small DeepSeek-V3.1-Chat & 37/671B & 35.9 & 22.3 & 33.2 & 22.8 & 28.8 & 44.5 & 30.7 & 45.3 & 38.7 & 23.1 & 29.2 & 46.3 & 31.8 & 47.6 & 35.9 & 21.0 & 27.0 & 43.3 & 29.0 & 44.2 \\
      \rowcolor{OpenColor} \small DeepSeek-V3.2-Think & 37/671B & 39.1 & 25.1 & 36.1 & 25.6 & 31.3 & 48.5 & 33.2 & 49.3 & 42.1 & 26.0 & 31.8 & 50.5 & 34.4 & 51.9 & 39.1 & 23.7 & 29.4 & 47.3 & 31.4 & 48.2 \\
      \rowcolor{OpenColor} \small DeepSeek-V3.2-Chat & 37/671B & 36.4 & 19.4 & 32.4 & 21.1 & 24.9 & 51.7 & 28.2 & 52.9 & 39.5 & 19.3 & 22.5 & 53.0 & 26.7 & 55.6 & 37.3 & 17.9 & 22.4 & 49.5 & 25.8 & 51.4 \\
      \rowcolor{OpenColor} \small Seed-Coder-8B-Instruct & 8B & 25.1 & 9.1 & 20.6 & 9.7 & 13.7 & 38.8 & 16.3 & 42.5 & 28.3 & 9.2 & 12.8 & 39.6 & 16.7 & 44.2 & 26.3 & 8.5 & 13.0 & 36.1 & 16.8 & 39.4 \\
      \rowcolor{OpenColor} \small Seed-OSS-36B-Instruct & 36B & 1.9 & 0.1 & 1.6 & 0.2 & 1.1 & 3.3 & 1.1 & 3.4 & 2.1 & 0.2 & 1.3 & 3.0 & 1.0 & 3.0 & 2.0 & 0.0 & 1.1 & 3.3 & 0.7 & 2.9 \\
      \midrule
      \rowcolor{OpenColor} \small Triton-SFT-32B & 32B &60.5 &47.6 &58.2 &45.2 &49.2 &67.0 &50.1 &66.5 &65.5 &53.1 &55.5 &75.0 &56.5 &75.0 &57.8 &45.0 &48.5 &66.8 &49.2 &66.7 \\
      \rowcolor{OpenColor} \small Triton-ORPO-32B & 32B &63.4 &53.2 &62.9 &53.1 &52.8 &72.1 &54.2 &72.3 &66.2 &55.5 &56.2 &75.0 &58.1 &75.3 &61.2 &51.0 &50.8 &70.5 &52.4 &71.1 \\
      \rowcolor{OpenColor} \small Triton-GRPO-32B & 32B &\bf70.1 &\bf58.7 &\bf69.9 &\bf58.2 &\bf59.6 &\bf79.5 &\bf60.7 &\bf79.8 &\bf72.5 &\bf60.2 &\bf61.1 &\bf82.6 &\bf62.6 &\bf83.8 &\bf67.8 &\bf52.1 &\bf57.7 &\bf77.7 &\bf57.9 &\bf78.7 \\
      \bottomrule
    \end{tabular}%
  }%
    \caption{Main results on the Mind2Web benchmark. We report the Composite Score and Step Success Rate (SR) across three generalization splits. \textbf{Triton-GRPO-32B} significantly outperforms all open-source baselines (including the 480B model) and surpasses top-tier proprietary models like GPT-4.5 and Claude-4.5 in Step SR, demonstrating robust generalization despite its smaller parameter size.}
  \label{tab:main_results}
  \vspace{-10pt}
\end{table*}

\section{Experiment}

\label{sec:experiments}

\subsection{Experimental Setup}

\paragraph{Datasets.}
Our experimental setup distinguishes between the data used for instruction tuning and the benchmarks employed for evaluation.

\noindent\textbf{Training Data.} 
To train our model, we utilize the full \textsc{Triton} dataset constructed in Section~\ref{sec:dataset} (approx. 590k instances). This incorporates both the discriminative trajectories mined from \textsc{Mind2Web}'s training split and the synthetic visual-grounding data from \textsc{WebSight}, ensuring the agent is exposed to diverse DOM structures and rejection scenarios before evaluation.

\noindent\textbf{Evaluation Benchmarks.} 
We evaluate our framework on the \textsc{Mind2Web} benchmark~\cite{deng2023mind2web}, which comprises over 1,000 tasks across diverse domains (see Table~\ref{tab:mind2web_stats}). Following standard protocols, we assess generalization across three splits with increasing distribution shifts: \textbf{(1) Cross-Task}: Unseen instructions on familiar websites; \textbf{(2) Cross-Website}: Unseen websites within known domains; and \textbf{(3) Cross-Domain}: Unseen websites in entirely new domains. The latter represents the most challenging setting, requiring robust structural generalization.

\begin{figure}[t]
    \centering
    \includegraphics[width=\linewidth]{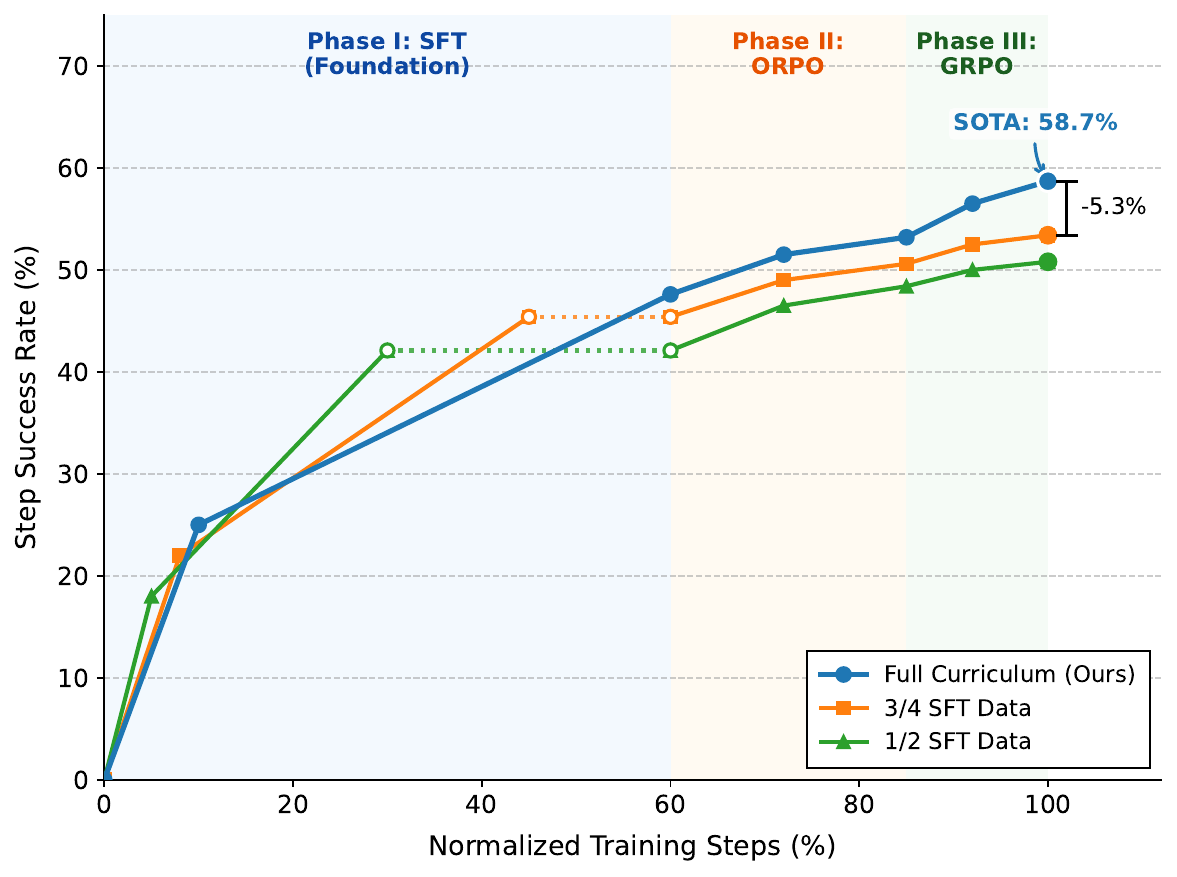}
\caption{Training dynamics analysis. Performance trajectories of the Full Curriculum versus variants with insufficient SFT foundations. Shaded areas denote alignment stages. The persistent gap confirms that ORPO and GRPO act as behavioral refiners rather than knowledge injectors, and thus cannot recover from a weak SFT baseline.}
    \label{fig:training_dynamics}
    \vspace{-15pt}
\end{figure}

\paragraph{Evaluation Metrics.}
We adopt the standard Mind2Web metrics~\cite{deng2023mind2web}: Element Accuracy (Ele. Acc) for selection correctness, Operation F1 (Op. F1) for text generation, and Step Success Rate (Step SR), which requires both to be correct. Additionally, we calculate a \textbf{Composite Score} averaging step-level ($\mu$) and task-level ($M$) performance:
\begin{equation}
    \text{Score} = \frac{1}{4} (\text{Acc}_{\mu} + \text{F1}_{\mu} + \text{Acc}_{M} + \text{F1}_{M})
\end{equation}

\paragraph{Implementation Details.}
We initialize our backbone with \textsc{Qwen2.5-Coder-32B-Instruct}~\cite{hui2024qwen2}, employing Llama Factory~\cite{zheng2024llamafactory} for supervised/preference alignment and VERL~\cite{verl2024} for RL training. Experiments are conducted on 8 NVIDIA H200 GPUs using the AdamW optimizer with cosine decay. The training curriculum proceeds as follows: (1) \textbf{SFT}: 3 epochs, lr $5\text{e-}5$, batch size 128, and seq length 8192; (2) \textbf{ORPO}: lr $5\text{e-}6$, batch size 64, and $\lambda=0.1$; (3) \textbf{GRPO}: lr $1\text{e-}6$, group size $G=5$, and $\beta_{\text{KL}}=0.001$. We employ greedy decoding for all evaluations. Comprehensive hyperparameters are detailed in Appendix~\ref{app:implementation}.

%%%%%%%%%%%%%%图表
\begin{table*}[!t]
\centering
\setlength{\tabcolsep}{4pt}
\small
\resizebox{\linewidth}{!}{
\begin{tabular}{l|ccc|cc|cc|cc|cc|cc}
    \toprule
    \multirow{2}{*}{\textbf{Model Variant}}
    & \multicolumn{3}{c|}{\textbf{SFT Triton Data}} & \multicolumn{2}{c|}{\textbf{RL Setting}}
    & \multicolumn{2}{c|}{\textbf{Overall }} 
    & \multicolumn{2}{c|}{\textbf{Cross-Domain }} 
    & \multicolumn{2}{c}{\textbf{Cross-Task }} 
    & \multicolumn{2}{c}{\textbf{Cross-Website }} \\
    & Mind &Dis & Vis & ORPO & GRPO
    & Score & Step SR 
    & Score & Step SR 
    & Score & Step SR 
    & Score & Step SR \\
    \midrule
    Qwen2.5-Coder-32B-Instruct  & \ding{55} & \ding{55} & \ding{55} & \ding{55} & \ding{55} & 28.7   & 14.6   & 30.1   & 15.5   & 29.8   & 16.1   & 26.2&27.9 \\
    + Mind& \cellcolor{waymollblue}\checkmark  & \ding{55}  & \ding{55} & \ding{55} & \ding{55} & 55.1 & 36.3 & 55.1  & 36.3  & 57.5 & 38.8   & 52.8 & 33.8  \\
    + Dis&  \ding{55} & \cellcolor{waymollblue}\checkmark  & \ding{55} & \ding{55} & \ding{55} & 60.7 & 42.0 & 59.8 & 41.9 & 64.8 & 45.6 & 57.7 & 38.6 \\
    + Vis&  \ding{55}  & \ding{55} &\cellcolor{waymollblue}\checkmark & \ding{55} & \ding{55} &  39.6& 21.5 &39.3 & 21.2  & 40.9 & 22.5   & 38.6 & 20.8  \\
    +Dis + Vis&  \ding{55}  & \cellcolor{waymollblue}\checkmark &\cellcolor{waymollblue}\checkmark & \ding{55} & \ding{55} & 60.5 & 47.6 & 58.2 & 45.2 & 65.5 & 53.1 & 57.8 & 45.0  \\
    +Dis + Vis +ORPO &  \ding{55}  & \cellcolor{waymollblue}\checkmark &\cellcolor{waymollblue}\checkmark & \cellcolor{waymollblue}\checkmark & \ding{55} & 63.4 & 53.2 & 62.9 & 53.1 & 66.2 & 55.5 & 61.2 & 51.0  \\
    \textbf{Triton-GRPO-32B (Ours)} &  \ding{55}  & \cellcolor{waymollblue}\checkmark &\cellcolor{waymollblue}\checkmark & \cellcolor{waymollblue}\checkmark & \cellcolor{waymollblue}\checkmark & \textbf{70.1} & \textbf{58.7} & \textbf{69.9} & \textbf{58.2} & \textbf{72.5} & \textbf{60.2} & \textbf{67.8} & \textbf{52.1}   \\
    \bottomrule
\end{tabular}
}
\vspace{-5pt}
\caption{Component-wise ablation study on the Mind2Web development set. We incrementally validate the effectiveness of our data construction strategies (Mind vs. Dis vs. Vis) and training stages (SFT $\rightarrow$ ORPO $\rightarrow$ GRPO). \textbf{Mind}: Original training set; \textbf{Dis}: Discriminative Trajectory Mining data; \textbf{Vis}: Synthetic Visual-Structural Grounding data. Note that \textbf{Triton-GRPO-32B}  combines all strategies to achieve optimal performance.}
\vspace{-15pt}
\label{tab:ablation_detailed}
\end{table*}

\paragraph{Baselines.}
We compare against representative SOTA models: \textbf{Open-Source} baselines include the Qwen series~\cite{qwen3,qwen3coder,qwen2.5}, DeepSeek family~\cite{deepseekv3,deepseekr1}, and Llama-3.1~\cite{llama3}, augmented by code-specialized models (e.g., OpenCoder~\cite{opencoder2024}, Seed-Coder~\cite{seed2025seedcoderletcodemodel}). \textbf{Closed-Source} baselines cover proprietary frontiers: GPT-4.5/o4-mini~\cite{openai2025gpt45}, Gemini 2.5~\cite{google2025gemini25}, Claude 4.5~\cite{anthropic2025claude}, and Grok-3~\cite{xai2025grok3}.

\subsection{Main Results}
\label{sec:main_results}

As shown in Table~\ref{tab:main_results}, \textbf{Triton-GRPO-32B} establishes a new SOTA with \textbf{58.7\%} Step SR, outperforming GPT-4.5 and Claude-4.5 by $>16\%$. This underscores three insights: \textbf{1) Efficiency:} Our 32B model doubles the performance of the 671B DeepSeek-V3 (25.2\%), proving that discriminative data outweighs raw scale in noisy environments. \textbf{2) Consistency:} GRPO significantly narrows the gap between Element Accuracy and Step SR, effectively curbing error propagation in long horizons. \textbf{3) Generalization:} On the unseen \textit{Cross-Domain} split, we boost the base model's performance from 15.5\% to 58.2\%. This confirms that the structural variance introduced by our synthetic data successfully equips the agent with robust transfer capabilities across novel domains.

\subsection{Ablation Study}
Table~\ref{tab:ablation_detailed} isolates the impact of our design choices:
\textbf{1) Mining Hard Negatives:} Discriminative mining (``+Dis'') outperforms naive training (``+Mind'') by \textbf{+5.7\%} (36.3\% $\rightarrow$ 42.0\%), proving that explicit boundary learning is superior to standard behavioral cloning.
\textbf{2) Structural Regularization:} While synthetic data (``+Vis'') lacks complex intents on its own (21.5\% SR), it significantly boosts Cross-Domain generalization (41.9\% $\rightarrow$ 45.2\%) when combined with real data, preventing layout overfitting.
\textbf{3) Alignment Synergy:} Our curriculum yields additive gains: ORPO improves the SFT baseline by \textbf{+5.6\%} (reducing hallucinations), and GRPO adds \textbf{+5.5\%} (ensuring consistency), culminating in a peak composite score of \textbf{70.1}.

\subsection{Dynamics Progressive Training Analysis.}
Figure~\ref{fig:training_dynamics} visualizes the evolution of Step Success Rate across the three curriculum stages.

\noindent\textbf{Stage-wise Performance Gains.} 
The \textit{Full Curriculum} (blue line) exhibits a consistent upward trajectory, validating the additive nature of our pipeline. SFT establishes a solid foundation (47.6\%), while ORPO provides a sharp discriminative boost (+5.6\%) by penalizing incorrect actions. Crucially, GRPO further refines consistency (+5.5\%) to reach the peak performance of 58.7\%. This distinct stepwise improvement suggests that \textit{discrimination} (knowing what not to do) and \textit{consistency} (maintaining performance over long horizons) are orthogonal capabilities that can be optimized sequentially.

\noindent\textbf{The Primacy of SFT Foundations.} 
Comparison with data-scarce baselines reveals the fundamental limits of post-training alignment. The ``1/2 SFT'' (green) and ``3/4 SFT'' (orange) variants, despite undergoing the identical ORPO and GRPO alignment stages, plateau significantly lower at 50.8\% and 53.4\%, respectively. Even with advanced reinforcement learning, the model cannot \textbf{compensate for} the lack of fundamental domain knowledge missed during the truncated SFT phase. This confirms our hypothesis that high-quality SFT acts as a prerequisite \textit{knowledge injector}, while ORPO and GRPO serve as \textit{behavioral regularizers}—they can steer the model's capabilities but cannot create capabilities that were never learned.

\subsection{Sensitivity of Discriminative Hyperparameters.}

\begin{figure}[t]
    \centering
    % 调整图片宽度，\linewidth 表示占满单栏宽度
    % 如果觉得图太大，可以改成 0.9\linewidth
    \includegraphics[width=\linewidth]{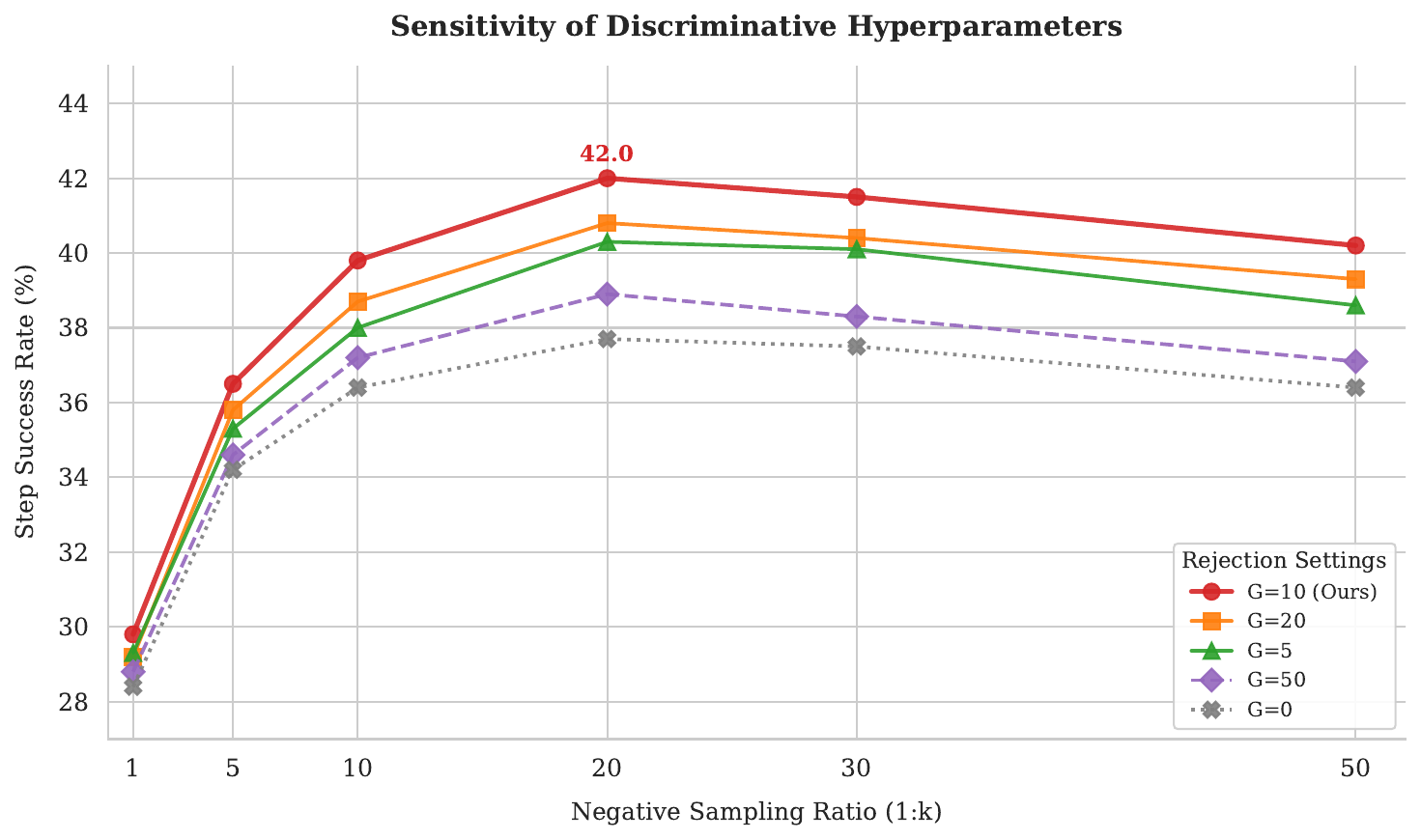} 
    
\caption{Hyperparameter sensitivity. Step Success Rate plotted against negative pool size ($K$) and rejection volume ($G$). The distinct peak at $K=20, G=10$ (Red line) indicates an optimal trade-off, demonstrating that performance degrades under both sparse discrimination signals ($K=1$) and excessive noise ($G=50$).}
    \label{fig:sensitivity}
    \vspace{-12pt}
\end{figure}

Figure~\ref{fig:sensitivity} analyzes the sensitivity of hard negative pool size ($K$) and rejection volume ($G$). We observe an inverted U-shaped trend for $K$, where a moderate pool ($K=20$) sharpens discrimination boundaries, while excessive scaling ($K=50$) overwhelms the model with noise. For rejection ($G$), the moderate setting ($G=10$) consistently outperforms aggressive strategies ($G=50$), which induce false negatives by making the policy overly conservative. Thus, we select $K=20$ and $G=10$ as the optimal configuration, balancing the trade-off between precision (hallucination suppression) and recall (target identification).

\subsection{Impact of Synthetic Data Scale.}
\begin{figure}[t]
    \centering
    \includegraphics[width=\linewidth]{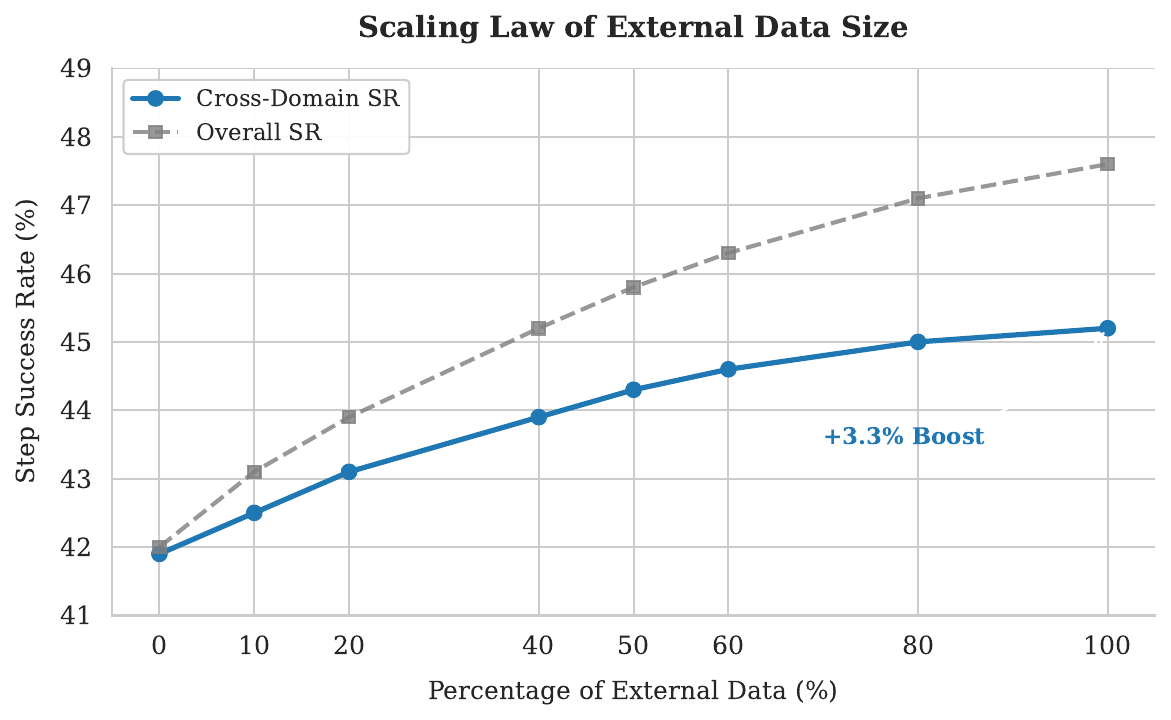}
\caption{Scaling analysis of synthetic data. We measure performance across varying data sizes (0--29k). The observed monotonic growth, particularly the \textbf{+3.3\%} boost in Cross-Domain (Blue), confirms that exposing the agent to diverse synthetic layouts serves as an effective structural regularizer for unseen websites.}
    \label{fig:scaling_law}
    \vspace{-12pt}
\end{figure}

Figure~\ref{fig:scaling_law} illustrates the trajectory as we scale the \textit{Synthetic Visual-Structural Grounding} data from 0\% to 100\% (approx. 29k samples). We observe a strictly monotonic, linear-like growth in both Overall and Cross-Domain performance without saturation, indicating that the model is far from its capacity ceiling. Notably, the \textit{Cross-Domain SR} achieves a robust \textbf{+3.3\% boost} (41.9\% $\rightarrow$ 45.2\%), empirically verifying that exposing the agent to diverse, auto-generated layouts serves as an effective structural regularizer to mitigate overfitting and enhance generalization to unseen domains.

%%%%%%%%%%%%%%图表

%%%%%%%%%%%%%%图表

\section{Related Works}

\paragraph{LLM-based Web Agent.}
The transition from simulated environments like MiniWoB++~\cite{liu2018reinforcement} to real-world web navigation has been accelerated by benchmarks such as Mind2Web~\cite{deng2023mind2web} and WebArena~\cite{zhou2023webarena}. 
Existing approaches can be broadly categorized into multimodal and text-based agents. 
Multimodal agents, such as SeeAct~\cite{zheng2024seeact} and CogAgent~\cite{hong2024cogagent}, leverage Visual Language Models (VLMs) to process screenshots, achieving high grounding accuracy but suffering from high latency. 
Text-based agents, relying on HTML DOM trees, offer a more efficient alternative. 
Recent works have focused on improving text-based reasoning through sophisticated prompting strategies, such as Chain-of-Thought (CoT)~\cite{wei2022chain} and tree-search planning~\cite{koh2024tree}. 
However, most prior research concentrates on architectural modifications or inference-time search, largely overlooking the critical role of data curriculum and discriminative training in enhancing robustness.

\paragraph{Data Synthesis for Code and Web.}
Synthetic data generation has become a paradigm for scaling instruction tuning, particularly in the code domain. 
Methods like Magicoder~\cite{wei2024magicoder} utilize LLMs to synthesize diverse coding problems and solutions, significantly improving performance. 
In the web domain, the WebSight dataset~\cite{laurent2024websight} provides large-scale screenshot-code pairs, primarily for converting designs to code. 
Recent efforts have attempted to repurpose such data for agent training; for instance, Auto-UI~\cite{kim2024autoui} generates synthetic instructions for mobile UIs. 
Our work extends this line of research by introducing a \textit{Dual-Agent Consensus} pipeline, ensuring that synthesized web navigation data is not only diverse but also strictly verifiable, addressing the hallucination issues common in naive synthesis.

\paragraph{Preference Alignment and Policy Optimization.}
Beyond Supervised Fine-Tuning (SFT), aligning LLMs with human preferences has proven essential for reducing errors. 
Reinforcement Learning from Human Feedback (RLHF)~\cite{ouyang2022training} is the standard approach but requires complex reward modeling. 
Direct Preference Optimization (DPO)~\cite{rafailov2024direct} simplifies this by optimizing the policy directly on preference pairs. 
More recently, Odds Ratio Preference Optimization (ORPO)~\cite{hong2024orpo} has been proposed to incorporate preference alignment directly into SFT, while Group Relative Policy Optimization (GRPO)~\cite{shao2024deepseekmath} focuses on enhancing reasoning capabilities by optimizing over groups of outputs without a critic model. 
To the best of our knowledge, our work is the first to systematically integrate these advanced alignment techniques (ORPO and GRPO) into a unified curriculum for web agents, specifically tailoring them to address the challenges of noisy HTML discrimination and long-horizon consistency.

\section{Conclusion}
This work bridges the gap between generalist LLMs and precise web navigation through the \textbf{Triton} dataset and a progressive curriculum. Through discriminative mining and synthetic grounding, we equip a compact 32B model with the ability to handle noisy DOMs and unseen domains. Empirical results on Mind2Web confirm that Triton-GRPO-32B outperforms massive proprietary models (e.g., GPT-4.5), validating a fundamental shift in agent training: superior performance stems not from parameter scale, but from the synergy of high-quality discriminative data and consistent policy alignment.

\section*{Limitations}
Despite the promising results, our framework encounters specific constraints. 
Primarily, Triton-GRPO-32B operates exclusively on HTML text representations; while synthetic data mimics structural layouts, the lack of a native visual encoder precludes the perception of pixel-level cues, such as color-coded status indicators or spatial occlusions. 
Additionally, our evaluation relies on the Mind2Web benchmark, which, being composed of static snapshots, may not fully capture the real-time dynamics of live browsing, including network latency, pop-ups, and CAPTCHAs. 
Finally, the integration of Group Relative Policy Optimization (GRPO) inevitably increases computational overhead compared to standard supervision, necessitating multiple trajectory rollouts for effective baseline estimation.

\section*{Ethics Statement}
This research adheres to ethical guidelines for AI development. We aim to enhance the capabilities of large language models (LLMs) while acknowledging potential risks such as bias, misuse, and privacy concerns. To mitigate these, we advocate for transparency, rigorous bias testing, robust security measures, and human oversight in AI applications. Our goal is to contribute positively to the field and to encourage responsible AI development and deployment.

\section*{Acknowledgments}
This work was supported in part by National Natural Science Foundation of China (No. 62506030), Beijing Natural Science Foundation (No. L242021), Young Elite Scientists Sponsorship Program of the Beijing High Innovation Plan (No. 20250938) and the Fundamental Research Funds for the Central Universitie (Grant No. GW2025-19) and supported by State Key Laboratory of Complex \& Critical Software Environment (Grant No. SKLCCSE-2025ZX-26).

% \input{latex/custom.bbl}
\bibliography{custom}

\appendix
% \onecolumn

\clearpage

\section{Data Construction Details}
\label{app:data_construction}

In this section, we provide detailed specifications for the dataset construction pipeline, including the mathematical formulation for hard negative mining and the automated synthesis workflow.

\subsection{Discriminative Trajectory Mining}
\label{app:mining_details}

To address the lack of negative constraints in standard behavioral cloning, we employ a \textit{Topological Hard Negative Mining} strategy. The core objective is to identify DOM elements that are structurally and semantically indistinguishable from the target element $e^+$ for a naive agent, thereby forcing the model to learn fine-grained discrimination.

\paragraph{Similarity Metric.}
We define a hybrid similarity function $S(e_i, e^+)$ between a candidate element $e_i$ and the ground truth target $e^+$. This function explicitly combines topological structure and semantic attributes:

\begin{equation}
    S(e_i, e^+) = \lambda \cdot S_{\text{topo}}(e_i, e^+) + (1 - \lambda) \cdot S_{\text{attr}}(e_i, e^+)
\end{equation}

\noindent where:
\begin{itemize}
    \item \textbf{Topological Similarity ($S_{\text{topo}}$)}: We utilize the Tree Edit Distance (TED) to measure the cost of transforming the subtree rooted at $e_i$ to $e^+$. The score is normalized as:
    \begin{equation}
        S_{\text{topo}} = 1 - \frac{\text{TED}(e_i, e^+)}{\max(|e_i|, |e^+|)}
    \end{equation}
    where $|e|$ denotes the number of nodes in the subtree. This captures layout similarities (e.g., identical list items).
    
    \item \textbf{Attribute Similarity ($S_{\text{attr}}$)}: We compute the Jaccard similarity of the attribute sets (class names, IDs, aria-labels, and inner text tokens):
    \begin{equation}
        S_{\text{attr}} = \frac{|\mathcal{A}(e_i) \cap \mathcal{A}(e^+)|}{|\mathcal{A}(e_i) \cup \mathcal{A}(e^+)|}
    \end{equation}
\end{itemize}

Algorithm~\ref{alg:hard_negative} details the computational process. We traverse the DOM tree $\mathcal{T}$ to compute similarity scores for all interactive elements. The top-$K$ elements with the highest $S$ scores are retained as hard negatives. In our experiments, we set $\lambda=0.6$ to prioritize structural similarity and $K=20$.

\begin{algorithm}[h]
\caption{Topological Hard Negative Mining}
\label{alg:hard_negative}
\begin{algorithmic}[1]
\Require 
    DOM Tree $\mathcal{T}$, 
    Target Element $e^+$, 
    Pool Size $K$, 
    Weight $\lambda$
\Ensure 
    Set of Hard Negative Elements $\mathcal{H}$

\State $\mathcal{C} \leftarrow \emptyset$ \Comment{Initialize candidate list with scores}
\State $\mathcal{E}_{\text{all}} \leftarrow \text{ExtractInteractiveElements}(\mathcal{T})$

\For{each element $e_i \in \mathcal{E}_{\text{all}}$}
    \If{$e_i = e^+$} 
        \State \textbf{continue} 
    \EndIf
    
    \State \textcolor{gray}{\# 1. Compute Topological Similarity via Tree Edit Distance}
    \State $d_{\text{tree}} \leftarrow \text{TreeEditDistance}(e_i, e^+)$
    \State $s_{\text{topo}} \leftarrow 1 - \frac{d_{\text{tree}}}{\max(\text{Size}(e_i), \text{Size}(e^+))}$
    
    \State \textcolor{gray}{\# 2. Compute Semantic Attribute Similarity}
    \State $A_i \leftarrow \text{GetAttributes}(e_i)$ \Comment{e.g., class, id, role}
    \State $A_+ \leftarrow \text{GetAttributes}(e^+)$
    \State $s_{\text{attr}} \leftarrow \frac{|A_i \cap A_+|}{|A_i \cup A_+|}$ \Comment{Jaccard Index}
    
    \State \textcolor{gray}{\# 3. Weighted Combination}
    \State $s_{\text{total}} \leftarrow \lambda \cdot s_{\text{topo}} + (1-\lambda) \cdot s_{\text{attr}}$
    
    \State $\mathcal{C}.\text{append}((e_i, s_{\text{total}}))$
\EndFor

\State \textcolor{gray}{\# 4. Select Top-K Hardest Negatives}
\State $\text{Sort}(\mathcal{C}, \text{key}=s_{\text{total}}, \text{order}=\text{DESC})$
\State $\mathcal{H} \leftarrow \{e \mid (e, s) \in \mathcal{C}[0:K]\}$

\Return $\mathcal{H}$
\end{algorithmic}
\end{algorithm}

\begin{figure*}[h]
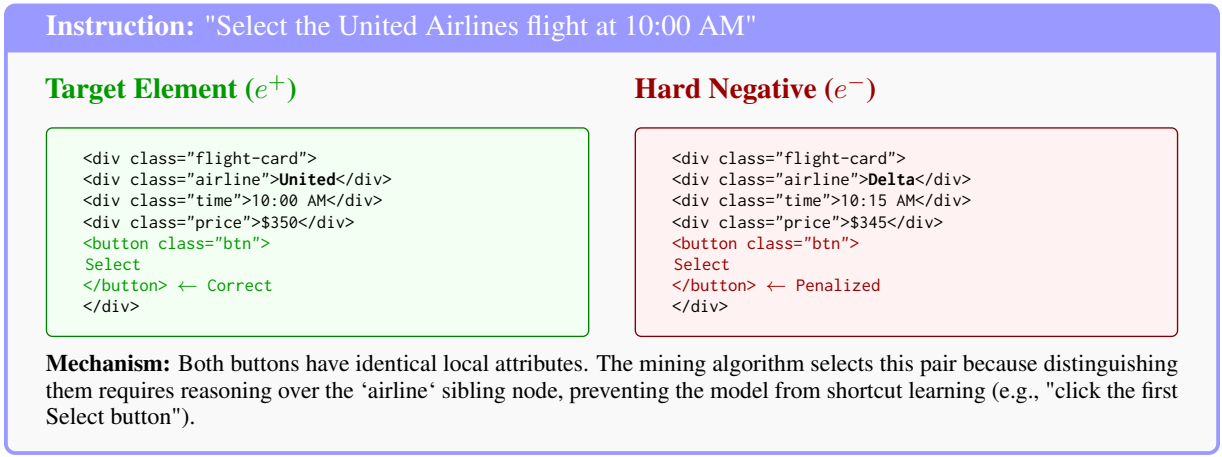

    \centering
    \begin{tcolorbox}[colback=gray!5, colframe=blue!40, title=\textbf{Instruction:} "Select the United Airlines flight at 10:00 AM"]
        
        % Left Side: Target
        \begin{minipage}{0.48\textwidth}
            \textbf{\textcolor{green!60!black}{Target Element ($e^+$)}}
            \begin{tcolorbox}[colback=green!5, colframe=green!50!black, boxrule=0.5pt, arc=2pt, fontupper=\scriptsize\ttfamily]
<div class="flight-card">\\
\ \ <div class="airline">\textbf{United}</div>\\
\ \ <div class="time">10:00 AM</div>\\
\ \ <div class="price">\$350</div>\\
\ \ \textcolor{green!60!black}{<button class="btn">}\\
\ \ \ \ \textcolor{green!60!black}{Select}\\
\ \ \textcolor{green!60!black}{</button>} \textcolor{green!60!black}{$\leftarrow$ Correct}\\
</div>
            \end{tcolorbox}
        \end{minipage}
        \hfill
        % Right Side: Negative
        \begin{minipage}{0.48\textwidth}
            \textbf{\textcolor{red!60!black}{Hard Negative ($e^-$)}}
            \begin{tcolorbox}[colback=red!5, colframe=red!50!black, boxrule=0.5pt, arc=2pt, fontupper=\scriptsize\ttfamily]
<div class="flight-card">\\
\ \ <div class="airline">\textbf{Delta}</div>\\
\ \ <div class="time">10:15 AM</div>\\
\ \ <div class="price">\$345</div>\\
\ \ \textcolor{red!60!black}{<button class="btn">}\\
\ \ \ \ \textcolor{red!60!black}{Select}\\
\ \ \textcolor{red!60!black}{</button>} \textcolor{red!60!black}{$\leftarrow$ Penalized}\\
</div>
            \end{tcolorbox}
        \end{minipage}
        
        \vspace{0.2cm}
        \footnotesize{\textbf{Mechanism:} Both buttons have identical local attributes. The mining algorithm selects this pair because distinguishing them requires reasoning over the `airline` sibling node, preventing the model from shortcut learning (e.g., "click the first Select button").}
    \end{tcolorbox}
    \caption{Visualizing Hard Negatives in a flight search list. The model must discern the correct button based on the associated airline context, despite high structural similarity.}
    \label{fig:hard_negative_flight}
\end{figure*}

\paragraph{Case Study 1: Hard Negative Mining (Flight Selection).}
We illustrate the challenge of distinguishing structurally identical elements in dense information lists using a flight booking scenario (Figure~\ref{fig:hard_negative_flight}). 

\textbf{Scenario:} The user issues the command: \textit{``Select the United Airlines flight departing at 10:00 AM''}. The search result page displays multiple flight cards with identical DOM trees (same nesting depth, same class names like \texttt{.flight-card}, \texttt{.btn-select}).
\begin{itemize}
    \item \textbf{Target ($e^+$)}: The "Select" button inside the United Airlines card.
    \item \textbf{Hard Negative ($e^-$)}: The "Select" button inside the Delta Airlines card (which appears immediately after).
\end{itemize}
As visualized below, the button element itself (`<button>Select</button>`) contains no distinguishing features. The agent must resolve the reference by attending to the \textit{sibling} element (`<div class="airline">`) or the \textit{parent} container's text. Our Topological Hard Negative Mining identifies the Delta flight button as the top ranked negative due to its minimal Tree Edit Distance, forcing the model to learn these non-local dependencies.

\begin{figure*}[h]
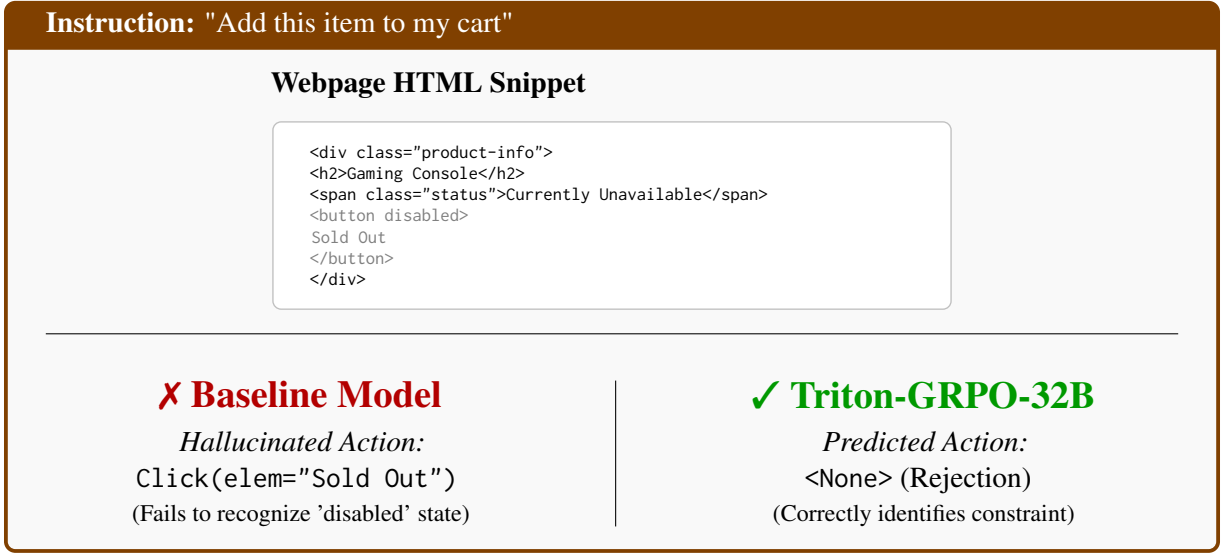

    \centering
    \begin{tcolorbox}[colback=gray!5, colframe=orange!50!black, title=\textbf{Instruction:} "Add this item to my cart"]
        
        % HTML Snippet Center
        \begin{center}
        \begin{minipage}{0.6\textwidth}
            \textbf{Webpage HTML Snippet}
            \begin{tcolorbox}[colback=white, colframe=gray!50, boxrule=0.5pt, fontupper=\scriptsize\ttfamily]
<div class="product-info">\\
\ \ <h2>Gaming Console</h2>\\
\ \ <span class="status">Currently Unavailable</span>\\
\ \ \textcolor{gray}{<button disabled>}\\
\ \ \ \ \textcolor{gray}{Sold Out}\\
\ \ \textcolor{gray}{</button>}\\
</div>
            \end{tcolorbox}
        \end{minipage}
        \end{center}
        
        \vspace{0.1cm}
        \noindent\rule{\textwidth}{0.4pt} % Separator line
        \vspace{0.1cm}

        % Comparison Bottom
        \begin{minipage}{0.45\textwidth}
            \centering
            \textcolor{red!70!black}{\textbf{\Large \ding{55} Baseline Model}} \\
            \vspace{0.1cm}
            \textit{Hallucinated Action:}\\
            \texttt{Click(elem="Sold Out")}
            \vspace{0.1cm}\\
            \footnotesize{(Fails to recognize 'disabled' state)}
        \end{minipage}
        \hfill
        \vline
        \hfill
        \begin{minipage}{0.45\textwidth}
            \centering
            \textcolor{green!60!black}{\textbf{\Large \ding{51} Triton-GRPO-32B}} \\
            \vspace{0.1cm}
            \textit{Predicted Action:}\\
            \texttt{<None>} (Rejection)
            \vspace{0.1cm}\\
            \footnotesize{(Correctly identifies constraint)}
        \end{minipage}

    \end{tcolorbox}
    \caption{Visualization of Counterfactual Rejection. While baseline models often force an action on disabled elements due to instruction bias, Triton correctly outputs a rejection token.}
    \label{fig:rejection_case}
\end{figure*}

\paragraph{Case Study 2: Counterfactual Rejection (Out-of-Stock).}
Standard SFT models suffer from hallucination when the requested action is impossible. We synthesize "Rejection" samples to address this. Figure~\ref{fig:rejection_case} demonstrates an "Action Mismatch" scenario.

\textbf{Scenario:} The user wants to \textit{``Add the item to cart''}, but the item is currently out of stock.
\begin{itemize}
    \item \textbf{Visual/HTML State:} The "Add to Cart" button is visually greyed out and has the `disabled` attribute, or is replaced by a "Sold Out" span.
    \item \textbf{Behavioral Cloning Trap:} A naive model, seeing the text "Sold Out" near a button-like shape, often hallucinates a click action because it overfits to the concept of "buying".
    \item \textbf{Triton Solution:} By training on counterfactuals where we intentionally disable valid elements and set the label to \texttt{<None>}, the agent learns to recognize the `disabled` attribute as a stop condition.
\end{itemize}

\subsection{Synthetic Visual-Structural Grounding}
\label{app:synthetic_pipeline}

To bridge the generalization gap between limited in-domain data and the diverse structural layouts of the open web, we introduce a \textit{Dual-Agent Consensus Pipeline}. This pipeline leverages the WebSight dataset, utilizing its screenshot-code pairs to synthesize high-quality instruction-trajectory samples.

\paragraph{Dual-Agent Consensus Mechanism.}
Automated data synthesis often suffers from two types of noise: (1) \textit{Unclear Instructions}, where the generated text is too vague, and (2) \textit{Hallucinated Grounding}, where the target element does not logically match the instruction. To mitigate this, we decouple the generation and verification processes:
\begin{itemize}
    \item \textbf{Generator ($M_G$)}: A large-scale coding model (Qwen3-Coder-480B-A35B-Instruct) that reasons over the HTML structure and screenshot to propose a plausible user intent and a target element.
    \item \textbf{Verifier ($M_V$)}: A strong instruction-following model (Qwen3-32B-Instruct) that acts as a critic. It receives \textit{only} the synthesized instruction and the raw HTML (blind to the Generator's choice) and attempts to predict the target element.
\end{itemize}
A sample is retained only if the Verifier's prediction strictly aligns with the Generator's label. This "consensus" ensures that the synthesized instruction is unambiguous and grounded in the DOM. The complete procedure is formalized in Algorithm~\ref{alg:consensus}.

\begin{algorithm}[h]
\caption{Dual-Agent Consensus Synthesis}
\label{alg:consensus}
\begin{algorithmic}[1]
\Require 
    WebSight Dataset $\mathcal{D}_{\text{raw}} = \{(H, S)\}_{i=1}^{N}$ (HTML, Screenshot),
    Generator $M_G$, 
    Verifier $M_V$, 
    Task Types $T = \{\text{Nav}, \text{Ret}, \text{Reas}\}$
\Ensure 
    Synthetic Dataset $\mathcal{D}_{\text{syn}}$

\State $\mathcal{D}_{\text{syn}} \leftarrow \emptyset$

\For{each pair $(H, S)$ in $\mathcal{D}_{\text{raw}}$}
    \State \textcolor{gray}{\# 1. Diversity-Driven Generation}
    \State $e_{\text{cand}} \leftarrow \text{RandomSelectInteractive}(H)$
    \State $type \sim \text{Uniform}(T)$ \Comment{Sample task cognitive level}
    
    \State \textcolor{gray}{\# Generator creates instruction based on element context}
    \State $I_{\text{syn}} \leftarrow M_G.\text{generate}(\text{prompt}(type, H, S, e_{\text{cand}}))$
    
    \If{$I_{\text{syn}}$ is INVALID or Empty}
        \State \textbf{continue}
    \EndIf

    \State \textcolor{gray}{\# 2. Blind Verification (Consensus Check)}
    \State \textcolor{gray}{\# Verifier predicts target given ONLY instruction and HTML}
    \State $e_{\text{pred}} \leftarrow M_V.\text{predict}(I_{\text{syn}}, H)$
    
    \State \textcolor{gray}{\# 3. Strict Consistency Filtering}
    \State $match \leftarrow \text{False}$
    \If{$e_{\text{pred}} == e_{\text{cand}}$} \Comment{Exact Match}
        \State $match \leftarrow \text{True}$
    \ElsIf{$\text{DOM\_Path\_Overlap}(e_{\text{pred}}, e_{\text{cand}}) > 0.9$} 
        \State \textcolor{gray}{\# Allow minor granularity diffs (e.g., button vs. inner span)}
        \State $match \leftarrow \text{True}$
    \EndIf
    
    \If{$match$ is \textbf{True}}
        \State $\mathcal{D}_{\text{syn}}.\text{append}(\{(H, I_{\text{syn}}, e_{\text{cand}})\})$
    \EndIf
\EndFor

\Return $\mathcal{D}_{\text{syn}}$
\end{algorithmic}
\end{algorithm}

\paragraph{Prompt Templates.}
To ensure the synthesized data covers diverse cognitive demands, we designed distinct prompt templates for the Generator. Additionally, the Verifier employs a strict grounding prompt to ensure alignment. The templates are detailed below.

\begin{figure*}[h!]
    \centering
    % GENERATOR PROMPTS
    \begin{tcolorbox}[colback=blue!5, colframe=blue!50!black, title=\textbf{Generator Agent Prompts ($M_G$)}]
        \scriptsize\ttfamily
        \textbf{[System System]} \\
        You are an expert web user specializing in creating realistic user interactions.
        
        \vspace{0.1cm}
        \textbf{[Shared Context]} \\
        \textbf{HTML Snippet:} \{HTML\_CONTEXT\} \\
        \textbf{Target Element:} \{TARGET\_ELEMENT\_HTML\}
        
        \vspace{0.1cm}
        \hrule
        \vspace{0.1cm}
        
        \textbf{[Task-Specific Instructions]}
        
        \textbf{Type 1: Navigation Intent} \\
        Generate a short, imperative command that directly operates on the target element. The command should be clear and action-oriented. \\
        \textit{Example Output: "Click the 'Sign Up' button at the top right."}
        
        \vspace{0.1cm}
        \textbf{Type 2: Information Retrieval} \\
        Generate a query that asks for specific information contained within the target element. The user is looking for content, not performing an action. \\
        \textit{Example Output: "What is the price of the Sony WH-1000XM4 headphones?"}
        
        \vspace{0.1cm}
        \textbf{Type 3: Reasoning Question} \\
        Generate a complex instruction that requires logical deduction or comparison with sibling elements to identify the target. Do not explicitly mention unique attributes (like IDs); describe the element by its relationship or condition. \\
        \textit{Example Output: "Select the flight that has the shortest duration among the options."}
    \end{tcolorbox}
    
    \vspace{0.2cm}
    
    % VERIFIER PROMPT
    \begin{tcolorbox}[colback=orange!5, colframe=orange!60!black, title=\textbf{Verifier Agent Prompt ($M_V$)}]
        \scriptsize\ttfamily
        \textbf{[System Message]} \\
        You are a precise web navigation assistant. Your goal is to identify the exact HTML element that matches the user's instruction.
        
        \vspace{0.1cm}
        \textbf{[User Input]} \\
        \textbf{Webpage HTML:} \\
        \{FULL\_HTML\_TREE\}
        
        \vspace{0.1cm}
        \textbf{User Instruction:} \\
        "\{SYNTHESIZED\_INSTRUCTION\}"
        
        \vspace{0.1cm}
        \textbf{[Constraint]} \\
        Analyze the HTML structure carefully. Return the \texttt{backend\_node\_id} of the element that best satisfies the instruction. If the instruction is ambiguous or the element is missing, output "None".
        
        \vspace{0.1cm}
        \textbf{[Response Format]} \\
        Thought: <Let's think step by step...> \\
        Target ID: <ID>
    \end{tcolorbox}
    \caption{Prompt templates used in the Dual-Agent Consensus pipeline. The Generator ($M_G$) is conditioned on the target to create diverse intents, while the Verifier ($M_V$) is blind to the target to ensure rigorous quality control.}
    \label{fig:prompt_templates}
\end{figure*}

\paragraph{Quality Control Example.}
To validate the necessity of the Verifier, we observed that the Generator occasionally produces \textit{visual hallucinations}. 
For instance, given a target element \texttt{<img src="logo.png">}, the Generator might produce \textit{"Click the red home icon"}, inferring color from the filename or screenshot. However, if the HTML lacks color attributes, the text-only Verifier ($M_V$) cannot ground "red" and correctly rejects the sample. This filtering step raised the grounding accuracy of our synthetic corpus from an initial 72\% to 96\% (as verified by human pilot study).

\section{Training Implementation Details}
\label{app:implementation}

\subsection{Input Processing \& Formatting}
\label{app:input_processing}

Raw HTML is notoriously verbose and token-inefficient. To adapt real-world websites to the context window of LLMs (8192 tokens), we employ a rigorous cleaning and formatting pipeline.

\paragraph{DOM Pruning and Cleaning.}
Before tokenization, the HTML DOM tree undergoes the following preprocessing steps:
\begin{enumerate}
    \item \textbf{Tag Removal}: We strip all non-visual and bulky tags, including \texttt{<script>}, \texttt{<style>}, \texttt{<svg>}, \texttt{<head>}, and \texttt{<meta>}.
    \item \textbf{Attribute Filtering}: To reduce noise, we retain only semantically relevant attributes: \texttt{class}, \texttt{id}, \texttt{type}, \texttt{name}, \texttt{aria-label}, \texttt{placeholder}, and \texttt{value}. Generic attributes (e.g., style strings, tracking codes) are discarded.
    \item \textbf{Text Truncation}: Long text nodes are truncated to the first 50 tokens to preserve the structural outline without exhausting the context window.
    \item \textbf{ID Injection}: Crucially, we inject a sequential numeric identifier (e.g., \texttt{backend\_node\_id}) into every interactive element. This allows the model to output a concise ID (e.g., "element=42") rather than generating a complex XPath.
\end{enumerate}

\paragraph{Multi-Task Input Format.}
We structure the input using the standard ChatML format. To support multi-step navigation, we also append the \textbf{Previous Action History} to the prompt. Figure~\ref{fig:input_format} illustrates the exact formatted input seen by the model.

\begin{figure*}[h]
    \centering
    \begin{tcolorbox}[colback=white, colframe=gray!60!black, title=\textbf{Formatted Model Input (ChatML style)}]
        \scriptsize\ttfamily
        \textcolor{blue!70!black}{<|im\_start|>system}\\
        You are a proficient web navigation agent. Given the HTML content and a user instruction, select the correct element and operation. 
        Output format: Element ID and Operation.\\
        \textcolor{blue!70!black}{<|im\_end|>}\\
        \textcolor{blue!70!black}{<|im\_start|>user}\\
        \textbf{Observation (Cleaned HTML):}\\
        \texttt{<html><body>...}\\
        \texttt{<div class="search-bar">}\\
        \texttt{\ \ <input id="42" placeholder="Search products...">}\\
        \texttt{\ \ <button id="43" aria-label="Search">Go</button>}\\
        \texttt{</div>}\\
        \texttt{...</body></html>}
        
        \vspace{0.1cm}
        \textbf{Previous Actions:}\\
        1. Type "iPhone 13" into element [15] (Search Box)
        
        \vspace{0.1cm}
        \textbf{Current Instruction:}\\
        "Click the search button to see results."\\
        \textcolor{blue!70!black}{<|im\_end|>}\\
        \textcolor{blue!70!black}{<|im\_start|>assistant}\\
        \textbf{Element:} 43\\
        \textbf{Operation:} Click\\
        \textcolor{blue!70!black}{<|im\_end|>}
    \end{tcolorbox}
    \caption{The flattened input representation. We inject explicit IDs (e.g., \texttt{id="42"}) into the HTML to enable precise referencing. The History block enables the model to maintain context across long horizons.}
    \label{fig:input_format}
\end{figure*}

\subsection{Alignment Details (ORPO \& GRPO)}
\label{app:alignment_details}

\paragraph{ORPO: Self-Synthesized Preference Pairs.}
Standard ORPO requires a dataset of $(y_w, y_l)$ pairs. Instead of using static negative samples, we construct \textit{dynamic, on-policy} negatives to target the SFT model's specific weaknesses.
\begin{enumerate}
    \item \textbf{Sampling}: For each instruction $x$ in the training set, we sample $N=5$ trajectories from the current SFT checkpoint $M_{\text{SFT}}$ using temperature $T=1.0$ to encourage diversity.
    \item \textbf{Winner Selection ($y_w$)}: The ground truth trajectory is always fixed as the winner.
    \item \textbf{Loser Selection ($y_l$)}: We select the "Hardest Loser" to penalize. Among the generated trajectories that are \textit{incorrect} (i.e., wrong element ID or wrong operation type), we select the one with the \textbf{highest log-probability}. 
\end{enumerate}
This selection strategy specifically targets the model's "blind spots"—answers that the model is confident in but are factually wrong (e.g., clicking a distinct button that looks similar to the target).

\paragraph{GRPO: Hierarchical Reward Structure.}
In the final stage, we employ Group Relative Policy Optimization (GRPO) to enforce consistency. Since web navigation involves strict formatting and precise text entry, a sparse binary reward is insufficient. 
We implement the hierarchical reward function defined in Eq. (3) of the main text. Table~\ref{tab:reward_breakdown} details the specific scalar values and conditions.

\begin{table}[h]
    \centering

    \vspace{0.2cm}
    \small
    \renewcommand{\arraystretch}{1.2}
    \begin{tabular}{l|c}
        \bottomrule
        \textbf{Hyperparameter} & \textbf{Value} \\
        \hline
        \multicolumn{2}{c}{\textit{General Configuration}} \\
        \hline
        Base Model & Qwen2.5-Coder-32B-Instruct \\
        Precision & bfloat16 \\
        Optimizer & AdamW \\
        Weight Decay & 0.01 \\
        LR Scheduler & Cosine \\
        Warmup Ratio & 0.03 \\
        Max Sequence Length & 8192 \\
        Hardware & 8 $\times$ NVIDIA H200 (141GB) \\
        \hline
        \multicolumn{2}{c}{\textit{Stage I: Foundation via SFT}} \\
        \hline
        Global Batch Size & 128 \\
        Learning Rate & $5 \times 10^{-5}$ \\
        Epochs & 3 \\
        Gradient Clipping & 1.0 \\
        Data Augmentation & None \\
        \hline
        \multicolumn{2}{c}{\textit{Stage II: Discrimination via ORPO}} \\
        \hline
        Global Batch Size & 64 \\
        Learning Rate & $5 \times 10^{-6}$ \\
        Epochs & 3 \\
        Odds Ratio Penalty ($\lambda$) & 0.1 \\
        Max Prompt Length & 4096 \\
        Max Completion Length & 1024 \\
        \hline
        \multicolumn{2}{c}{\textit{Stage III: Consistency via GRPO}} \\
        \hline
        Group Size ($G$) & 5 \\
        Learning Rate & $1 \times 10^{-6}$ \\
        KL Coefficient ($\beta_{\text{KL}}$) & 0.001 \\
        Rollout Batch Size & 32 \\
        Episodes per Step & 10 \\
        Discount Factor ($\gamma$) & 1.0 \\
        Sampling Temperature & 1.0 (Training) / 0.0 (Eval) \\
        \toprule
    \end{tabular}
    \caption{\textbf{Hyperparameter Settings.} We list the specific configurations for SFT, ORPO, and GRPO stages. Note that the learning rate decays according to a cosine schedule across all stages.}
    \label{tab:hyperparameters}
\end{table}

\begin{table*}[t]
    \centering
    \vspace{0.2cm}
    \small
    \renewcommand{\arraystretch}{1.3}
    \begin{tabular}{l|c|p{12cm}} 
        \toprule
        \textbf{Component} & \textbf{Value} & \textbf{Condition / Description} \\
        \hline
        \textbf{Format Reward} ($R_{\text{fmt}}$) & $+0.1$ & Awarded if the output strictly follows the JSON/ChatML structure (valid Element ID and Operation fields). \\
        \hline
        \textbf{Option Gate} ($\mathbb{I}_{\text{opt}}$) & $\{0, 1\}$ & Acts as a multiplier. $1$ if the predicted Element ID matches the ground truth, $0$ otherwise. \\
        \hline
        \textbf{Element Reward} ($R_{\text{opt}}$) & $+1.0$ & Major reward for correctly identifying the target DOM element. \\
        \hline
        \textbf{Arg F1 Score} ($R_{\text{F1}}$) & $[0, 1.0]$ & Token-level F1 score between predicted typed text (e.g., "iphone 13") and ground truth. Only calculated if $\mathbb{I}_{\text{opt}}=1$. \\
        \hline
        \textbf{Perfect Bonus} ($R_{\text{perf}}$) & $+1.0$ & Extra bonus if both Element and Operation arguments are perfectly correct ($F1=1.0$). \\
        \hline
        \textbf{Max Total Reward} & \textbf{3.1} & $0.1 + 1 \cdot (1.0 + 1.0 + 1.0)$ \\
        \bottomrule % 修正：底部用 bottomrule
    \end{tabular}
    \caption{\textbf{Hierarchical Reward Breakdown.} The reward function is designed as a cascade: the agent must first satisfy the format, then the element selection, and finally the operation arguments. If the element is wrong ($\mathbb{I}_{\text{opt}}=0$), subsequent rewards are zeroed out to prevent reward hacking.}
    \label{tab:reward_breakdown}
\end{table*}

\paragraph{Reward Calculation Examples.}
To illustrate, consider a ground truth action: \texttt{Type "Apple" into ID=42}.
\begin{itemize}
    \item \textbf{Case A (Perfect):} Output is \texttt{ID=42, Type "Apple"}.
    \\[0.1cm]
    $R = 0.1 (\text{fmt}) + 1.0 \cdot (1.0 (\text{opt}) + 1.0 (\text{f1}) + 1.0 (\text{perf})) = \mathbf{3.1}$.
    
    \item \textbf{Case B (Minor Typo):} Output is \texttt{ID=42, Type "App"}.
    \\[0.1cm]
    $R = 0.1 + 1.0 \cdot (1.0 + 0.5 (\text{partial f1}) + 0.0) = \mathbf{1.6}$. The agent is encouraged for getting the right element but penalized for the typo.
    
    \item \textbf{Case C (Wrong Element):} Output is \texttt{ID=99, Type "Apple"}.
    \\[0.1cm]
    $R = 0.1 + 0.0 \cdot (\dots) = \mathbf{0.1}$. Even though the text "Apple" is correct, the wrong element zeros out the rest. This enforces strict grounding.
\end{itemize}

\subsection{Hyperparameters}
\label{app:hyperparams}

We report the detailed hyperparameters used across the three stages of our Progressive Training Curriculum in Table~\ref{tab:hyperparameters}. All experiments were conducted on a cluster of 8 NVIDIA H200 GPUs. The backbone model is initialized from \texttt{Qwen2.5-Coder-32B-Instruct}.

\end{document}